\documentclass[12pt]{article}
\usepackage{amssymb,amsmath,cite,bm}
\usepackage[dvips]{graphicx,color}
\usepackage{psfrag,subfigure}
\begin{document}

\title{\bf Autonomous Robots for Active Removal of Orbital Debris}

\author{Farhad Aghili}

\date{}

\maketitle

\begin{abstract}
This paper presents a vision guidance and control method for autonomous robotic capture and stabilization of a tumbling orbital debris object in a time-critical manner. The method takes into account various operational and physical constraints, including ensuring a smooth capture, handling line-of-sight (LOS) obstructions of the target, and staying within the acceleration, force, and torque limits of the robot. Our approach involves the development of an optimal control framework for an eye-to-hand visual servoing method, which integrates two sequential sub-manoeuvres: a pre-capturing manoeuvre and a post-capturing manoeuvre, aimed at achieving the shortest possible capture time. Integrating both control strategies enables a seamless transition between them, allowing for real-time switching to the appropriate control system. Moreover, both controllers are adaptively tuned through vision feedback to account for the unknown dynamics of the target.  The integrated estimation and control architecture also facilitates fault detection and recovery of the visual feedback in situations where the feedback is temporarily obstructed. The experimental results demonstrate the successful execution of pre- and post-capturing operations on a tumbling and drifting target, despite multiple operational constraints and the presence of obstructed 3D vision data.
\end{abstract}

                        





\section{Introduction}

\begin{figure}
\centering \includegraphics[width=8cm]{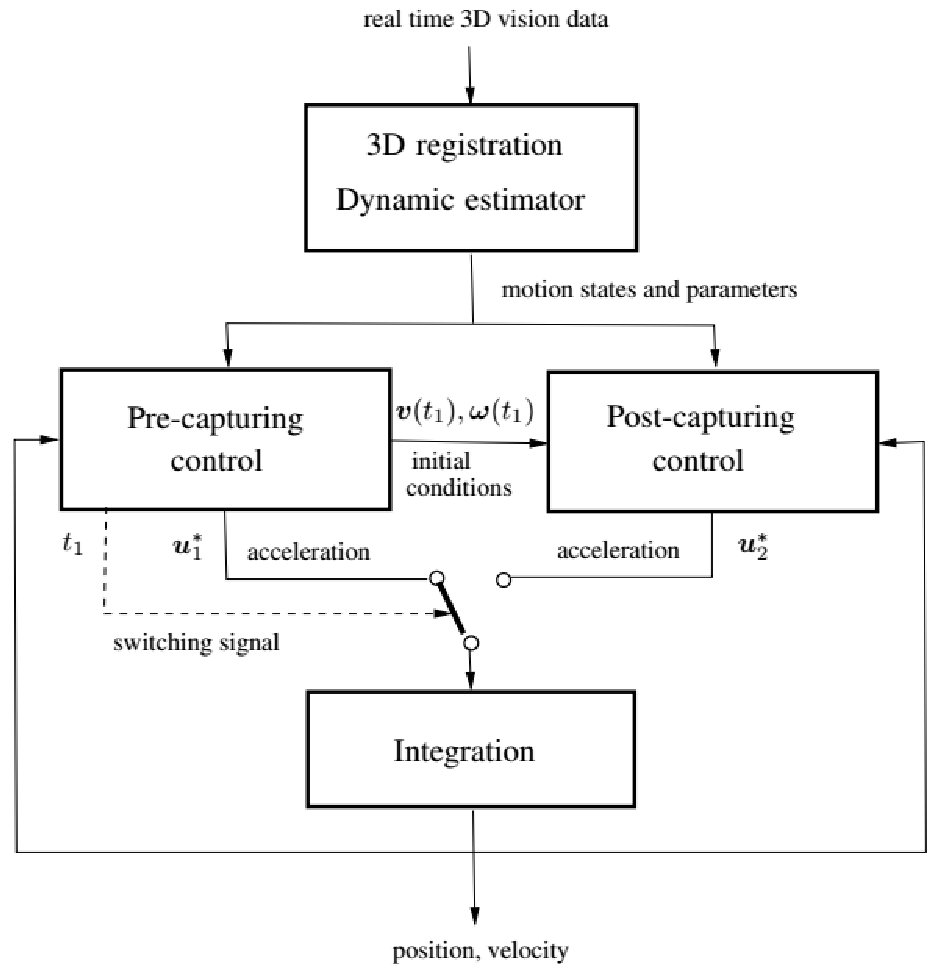}
\caption{Control architecture for autonomous sequential pre- and post-capturing of space objects.}\label{fig:control2}
\end{figure}

Autonomous servicing robotics encompasses a broad range of integrated technologies, including intelligent guidance and controls, vision systems, as well as specialized capturing end-effectors and tools. The application of autonomous robots for on-orbit servicing has opened up new opportunities for the commercial sector, national space agencies, and universities. Servicing operations cover a wide range of tasks, including maintenance and repair, rescue missions, refuelling, inspections, rendezvous and docking, as well as orbital debris removal~\cite{Aghili-2023,Papadopoulos-Moosavian-1994,Yoshida-Dimitrov-Nakanishi-2006,Aghili-2011k,Wang-Meng-2020,Aghili-Parsa-2007b,Wang-Huang-2015,Aghili-2013,Kang-Zhu-2021,Aghili-Parsa-2009b,Aghili-2020a}. All these robotic servicing mission concepts require an autonomous robotic arm to reliably capture a target space object with non-zero relative translational and rotational motions, subject to multiple constraints. Many of these target satellites are considered non-cooperative objects because they were not designed or built with the intention of being serviceable in the future. Moreover, these space objects often have tumbling motions due to non-functional attitude control systems, making robotic servicing of non-cooperative satellites extremely challenging. The space robot must first capture the tumbling satellite and then safely remove its angular momentum before executing subsequent repairing, rescuing, or de-orbiting operations. Therefore, we divide the robotic capture and stabilization task into two primitive robot operations: (i) pre-capturing manoeuvre and (ii) post-capturing manoeuvre, as illustrated schematically in Fig.~\ref{fig:control2}.

Despite significant progress made in the past two decades, vision-guided robotic systems still face many challenging problems. These challenges arise mainly due to the undependability of vision systems, environmental uncertainties, and multiple systems and operational constraints. A reliable vision-guided robotic system should be capable of adaptively tuning itself against inaccurate and potentially erroneous visual information, as well as uncertainties affecting the system performance. Since autonomous capture and stabilization of the client satellite is a time-critical operation, completing the entire operation as quickly as possible given the constraints is crucial. To achieve this objective, it is necessary to integrate sequential sub-manoeuvres associated with both the pre-capturing and post-capturing phases of the robot guidance problem in an optimal and seamless manner.

Despite the existence of various guidance and control strategies for robotic interception of moving objects, including vision-based motion estimation techniques described in \cite{Wee-Walker-1993,Papadopoulos-Moosavian-1994,Gregorio-Ahmadi-Buehler-1997,Croft-Fenton-Benhabib-1998,Aghili-Parsa-2007b,Rybus-Seweryn-2014,Chwa-Kang-2005,Mehrandezh-Sela-Fenton-Benhabib-2000,Aghili-2011k,Koivo-Houshangi-1991} and others, seamless robotic planning in both pre- and post-capturing phases, which satisfies time-criticality of the entire operation while handling multiple constraints in a reliable manner, still poses a significant challenge. In the literature, a number of optimal and non-optimal robot-motion planning and guidance techniques have been developed for interception of moving targets, including those presented in \cite{Sharifi-Wilson-1998,Mehrandezh-Sela-Fenton-Benhabib-2000,Chwa-Kang-2005,Wang-Guo-2017,Zong-Luo-2020,Rekleitis-2015,Aghili-Parsa-2008b}. Additionally, a planning and control methodology has been proposed in \cite{Rekleitis-2015} for manipulating passive objects by collaborating with orbital free-flying servicers in zero gravity. Various visual-tracking control approaches for space manipulators capturing target spacecraft in uncertain dynamics are presented in the literature. For instance, \cite{Aghili-2011k, Wang-Guo-2017} describe predictive visual servo kinematic control schemes for autonomous capture of non-cooperative space targets with unknown motion, while \cite{Zong-Luo-2020} present an optimal control method for space manipulators that saves on-board fuel and satisfies obstacle avoidance targets during rendezvous and capture. In \cite{Lampariello-Mishra-2018}, a tracking control method for grasping tumbling satellites is presented, which employs a visual servo for the approach phase and an online EKF estimator to account for modelling uncertainties \cite{Aghili-Kuryllo-Okouneva-English-2010a,Aghili-Parsa-2009,Aghili-Kuryllo-Okuneva-McTavish-2009,Aghili-2008c}.
A solution to the guidance problem of capturing a tumbling space object based on convex programming formulation is proposed in \cite{Virgili-Llop-2019}. This approach builds on the earlier work on optimal trajectory planning for rendezvous and proximity operation using non-convex keep-out-zone constraints as presented in \cite{Lu-Liu-2013}. Additionally, \cite{Zang-Zhang-2020} proposes a detumbling system that involves the robot and the target, where the target's energy is gradually dissipated through contact effects. Although recent surveys such as \cite{Papadopoulos-Aghili-2021, Moghaddam-Chhabra-2021,Aghili-2009a,Aghili-Parsa-Martin-2008a} cover various research works on robotic trajectory planning and capture in space, there is a notable lack of literature regarding a seamless control strategy for integrating both pre- and post-capturing phases and the corresponding end-to-end experimental validation.


This work  presents a seamless integration of two optimal control strategies for autonomously capturing and stabilizing a moving/tumbling satellite by utilizing 3D vision feedback~\cite{Aghili-2023}, see Fig.~\ref{fig:control2}. 
This work builds upon our earlier contributions in \cite{Aghili-2011k} by introducing adaptive and consistent optimal solutions for both the pre-capturing and post-capturing phases. This includes enabling a smooth transition between the two control strategies during pre- and post-capturing operation phases, as well as real-time switching to the appropriate control system. Furthermore, we introduce an innovative dynamics formulation that enables self-tuning of the trajectory planner not only in the pre-capturing phase but also in the post-capturing phase based on feedback from the vision system. The aim  is to improve the time-criticality, reliability, and adaptability of autonomous robots during proximity operations in space, with three main objectives: (i) developing an end-to-end time-optimal trajectory planning for the two sequential sub-manoeuvres while considering multiple operational and physical constraints to ensure time-criticality of autonomous operations, (ii) enhancing the reliability and robustness of the autonomous proximity operation by creating a fault-tolerant vision-guided system that can continuously operate even if the vision sensor generates erroneous information, and (iii) adapting the planning process to parametric uncertainties for improved adaptability. In order to achieve this goal, a hierarchical control system is developed for the autonomous robotic capture and stabilization of a target that has both translational and tumbling motions, despite various physical limitations, uncertainties, and temporary visual obstructions, in a time-critical manner. The system features adaptive deliberate planning and a seamless optimal trajectory plan for two sequentially occurring sub-manoeuvres, taking into account multiple operational and physical constraints to meet the time-critical demand of the autonomous proximity operation. To evaluate the performance and robustness of the proposed robot guidance and control strategy, experiments are conducted using a ground-based satellite simulator testbed \cite{Aghili-Dupuis-Piedboeuf-deCarufel-1999}.

\section{Modelling \& Motion Estimation using Occluded Vision Data} \label{sec:ICP}
\begin{figure}
\centering\includegraphics[width=8cm]{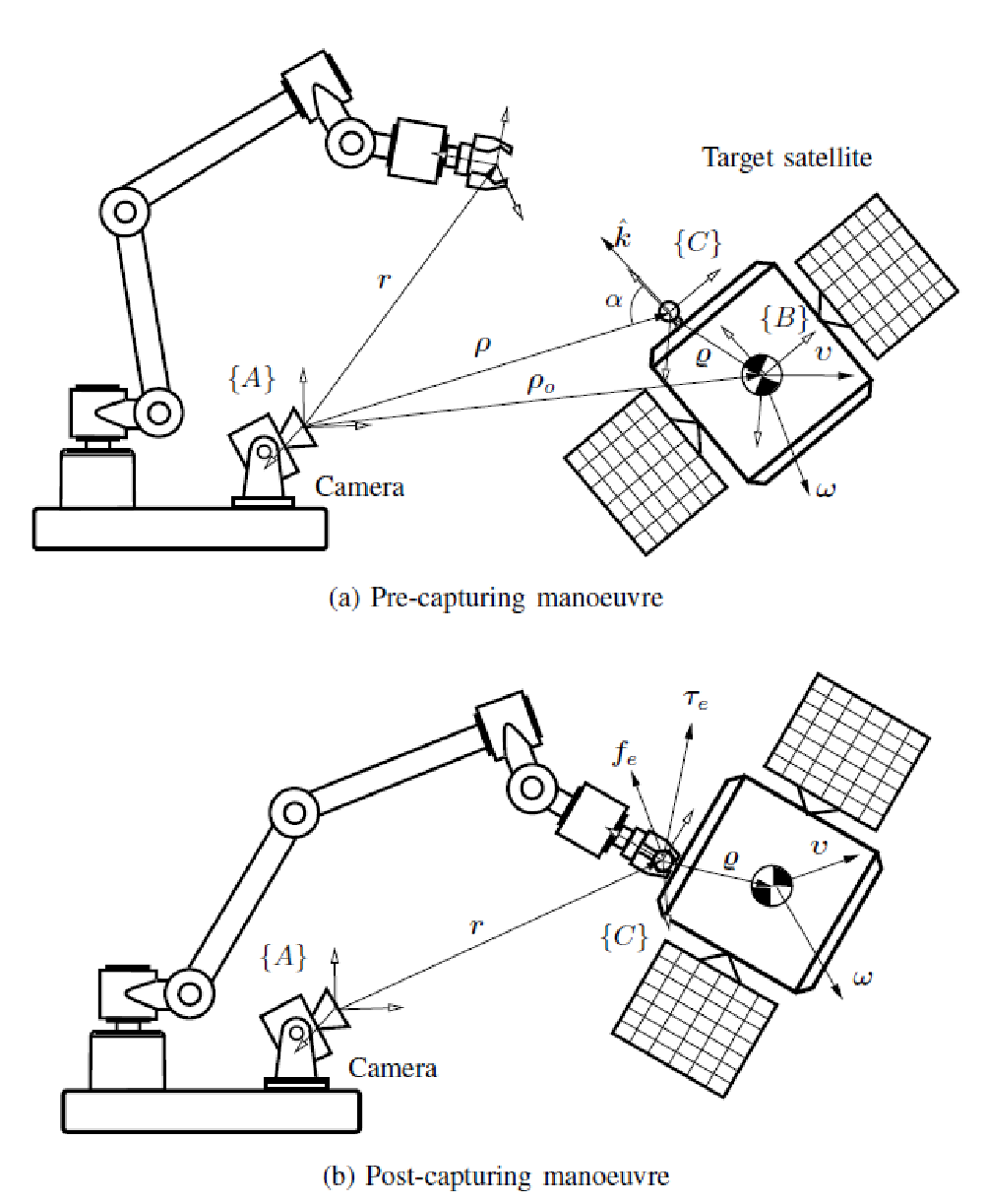}
 \caption{Vision-guided manipulator (eye-to-hand positioning of the
camera) and target during pre- and post-capturing manoeuvres.} \label{fig:robot_target}
\end{figure}

Fig.~\ref{fig:robot_target} depicts the coordinate frames used for a vision-guided manipulator system in the pre- and post-capture phases of a tumbling target satellite \cite{Aghili-2009d}. The camera coordinate frame is denoted as ${A}$, while coordinate frames ${B }$ and ${ C}$ are attached to the body of the target. The origin of frame ${B }$ coincides with the center-of-mass (CoM) of the target, while the origin of frame ${ C}$ is placed at a distance $\bm\varrho$ from the CoM, representing the location of the grasping fixture. We assume that frame $\{ B \}$ is aligned with principal axes of the body. The measurement of the pose (the position and attitude) of coordinate $\{{\cal C}\}$  with respect to the coordinate
frame $\{ A\}$ represented by variables $\bm\rho$ and unit quaternion $\bm\eta$. Suppose unit quaternions $\bm\mu$ and $\bm q$,  represent the orientations of
coordinates frames $\{ B \}$ respect to  $\{ C\}$  and $\{ B \}$ respect to $\{ A \}$, respectively. Then, quaternion $\bm\eta$ combines two orientations and thus we have
\begin{equation} \label{eq:muOtimesq}
\bm\eta = \bm\mu \otimes \bm q, \quad \mbox{where} \quad  \bm\mu \otimes = \mu_o \bm I + \bm\Omega(\bm\mu_v)
\end{equation}
is the quaternion product operator, $\bm\mu_v$ denotes the vector part of quaternion $\bm\mu$,   and
\begin{equation} \label{eq:Omega}
 \bm\Omega(\bm\mu_v)=\begin{bmatrix} -[\bm\mu_v
\times] & \bm\mu_v \\ - \bm\mu_v^T & 0 \end{bmatrix}.
\end{equation}
Notice that since the target rotates, quaternion $\bm\eta$ and $\bm q$ are time-varying variables whereas quaternion $\bm\mu$ is a constant. Also, one can infer from the
schematics in Fig.~\ref{fig:robot_target}.a that the following kinematics relationship holds
\begin{equation} \label{eq:rho}
\bm\rho = \bm\rho_o + \bm A(\bm q) \bm\varrho,
\end{equation}
where $\bm\rho_o$  denotes the location of the target's CoM that is given in the coordinate frame $\{ A  \}$, and the rotation matrix $\bm A(\bm q)$ as a function of
quaternion $\bm q$ is given by
\begin{equation} \label{eq:R}
\bm A(\bm q) =\bm I + 2 q_o [\bm q_v \times] + 2 [\bm q_v \times]^2,
\end{equation}
Here, the quaternion $\bm q=[\bm q_v^T \; q_o]^T$ is decomposed into the vector part, $\bm q_v$, and the scaler part, $q_o$, while the matrix form of the cross-product is denoted by $[\cdot \times]$.

Suppose $\bm\omega$ represents the target's angular velocity expressed in body-fixed frame $\{ B \}$. Then, the rotational and translational motions of the target in the pre-capturing phase can be described by
\begin{subequations} \label{eq:Euler-Newton}
\begin{align}
\bm I_c \dot{\bm\omega} & =\bm \omega \times \bm I_c \bm\omega  +  \bm\tau_{\rm dis} \\
m \ddot{\bm\rho}_o & = \bm f_{\rm dis} \\ \label{eq:dot_q}
\dot{\bm q} & = \frac{1}{2} \bm\Omega(\bm \omega) \bm q
\end{align}
\end{subequations}
where $\bm I_c =\mbox{diag}(I_{xx}, I_{yy},I_{zz})$ is the target inertia tensor in terms of the principal moments of inertia,  $\bm\tau_{\rm dis}$ and $\bm
f_{\rm dis}$ are small torque and force disturbances acting on the target satellite. In the following analysis, we will re-write the above dynamics equations in terms of a set of
identifiable inertia parameters. This is because \eqref{eq:Euler-Newton} is not an adequate formulation for dynamics identification problem requiring the minimum number of inertial parameters. Define the following non-dimensional inertia parameters:
\begin{equation} \label{eq:p}
\sigma_1 = \frac{I_{yy}- I_{zz}}{I_{xx}}, \quad
\sigma_2 = \frac{I_{zz}- I_{xx}}{I_{yy}}, \quad
\sigma_3 = \frac{I_{xx}- I_{yy}}{I_{zz}},
\end{equation}
The principal moments of inertia satisfy the following triangular inequalities
\begin{align} \notag
I_{xx}+ I_{yy} > I_{zz} \\ \notag
I_{yy}+ I_{zz} > I_{xx} \\ \label{eq_inequality}
I_{zz}+ I_{xx} > I_{yy}
\end{align}
From \eqref{eq:p} and \eqref{eq_inequality}, one can show by inspection that the following equality and inequality  constraints between the dimensionless parameters are in order
\begin{subequations}  \label{eq:sigma_bounds}
\begin{align} \label{eq:sigma_constraint}
& \sigma_1 + \sigma_2 + \sigma_3 + \sigma_1\sigma_2\sigma_3   =0, \\
& -1 < \sigma_i  < 1 \qquad \forall i=1,\cdots,3. \
\end{align}
\end{subequations}
The equality constraint \eqref{eq:sigma_constraint} implies that the dimensionless parameters are not independent of each other. Considering a compact set of two dimensionless inertia
parameters \begin{equation} \notag
\bm\sigma = \begin{bmatrix} \sigma_1 \\ \sigma_2 \end{bmatrix},
\end{equation}
one can obtain the third variable from \eqref{eq:sigma_constraint} as follow:
\begin{equation}
\sigma_3= - \frac{\sigma_1 + \sigma_2}{1 + \sigma_1 \sigma_2}
\end{equation}
We can also concisely express the set of inequalities \eqref{eq:sigma_bounds} as a vector inequality:
\begin{equation} \label{eq:Dsigma<1}
- \bm 1 < \bm\sigma < \bm 1
\end{equation}
where $\bm 1=[1\;\; 1]^T$ is the vector of one. Now, we are ready to express the Euler's rotation equations  in terms of the independent dimensionless parameters
$\bm\sigma$ as follows
\begin{subequations} \label{eq:target_dyn}
\begin{align} \label{eq:dot_omega}
\dot{\bm\omega} & = \bm\phi(\bm\omega, \bm\sigma) + \bm B(\bm\sigma)
\bm w_{\tau}, \\
\ddot{\bm\rho}_o & = \bm w_{f}.
\end{align}
Here, $\bm w_{\tau} = \bm\tau_{\rm dis}/{\rm tr}(\bm I_c)$ is the angular acceleration disturbance,   $\bm w_{f} = \bm f_{\rm dis}/m$ is the linear acceleration
disturbance,  ${\rm tr}(\cdot)$ is the trace operator, and
\begin{align} \notag
\bm B(\bm \sigma) & = \begin{bmatrix}\frac{\pi(\bm\sigma)}{1 - \sigma_2} & 0 & 0 \\
0 & \frac{\pi(\bm\sigma)}{1 + \sigma_1} & 0 \\
0 & 0 & \frac{\pi(\bm\sigma)}{1 + \sigma_1 \sigma_2} \end{bmatrix}, \\
\pi(\bm\sigma) & = 3 + \sigma_1\sigma_2 + \sigma_1 -\sigma_2, \\
\bm\phi(\bm\omega, \bm\sigma) &= \begin{bmatrix} \sigma_1 \omega_y \omega_z \\ \sigma_2 \omega_x \omega_z \\
-\frac{\sigma_1 + \sigma_2}{1 + \sigma_1 \sigma_2} \omega_x \omega_y  \end{bmatrix}.
\end{align}
\end{subequations}
We assume the angular and linear acceleration disturbances to be zero-mean noises with covariances $E[\bm w_{\tau} \bm w_{\tau}^T ] = \sigma_{\tau}^2 \bm I$ and $E[\bm
w_{f} \bm w_{f}^T ] = \sigma_{f}^2 \bm I$, where $E[\cdot]$ is the expected operator.

Consider the following state vector pertaining to both states  and the associated dynamic parameters
\begin{equation} \label{eq:x}
\bm x= \begin{bmatrix}\bm q_v \\ \bm\omega \\ \bm\rho_o \\ \dot{\bm\rho}_o \\ \bm\theta \end{bmatrix} \quad \mbox{where} \quad \bm\theta = \begin{bmatrix} \bm\sigma \\
\bm\varrho \\  \bm\mu_v \end{bmatrix}
\end{equation}
contains the constant parameters, i.e.,
\begin{equation}
\dot{\bm\theta} = \bm 0.
\end{equation}
Thus, the location of CoM, the inertia ratio, the orientation of the principal axes are assumed to be unknown.

Assuming a given surface model of the target and the availability of three-dimensional (3-D) point measurements through an active vision system, we proceed with the analysis. Let data set $\{\bm c_{1} \cdots \bm c_{m} \}$ represent the 3D points data acquired by scanning an object  at time $t$,  while the surface model of the same object is represented by model
set ${\cal M}$. Here, vector $\bm c_{i} \in\mathbb{R}^3$  represents the coordinate of $i$th single point from the point cloud.  For each point $\bm c_{i}$ from the data points set, one can find the  corresponding point  $\bm d_{i}\in\mathbb{R}^3$ on the surface model ${\cal M}$.
Note that vectors $c_i$s are expressed in frame in the Camera coordinate frame $\{ A \}$. Therefore, one should be able to populate the date set $\{\bm d_{1} \cdots \bm d_{m} \}$ representing  all corresponding points to the  data set $\{\bm c_{1} \cdots \bm c_{m} \}$ through an optimization process~\cite{Simon-Herbert-Kanade-1994}. Therefore, the instantaneous pose of the target, represented by translation vector $\bm\rho$ and quaternion $\bm q$, can be  written as a function of the point cloud set, i.e.,
\begin{equation} \label{eq:y}
\bm y(\bm c_1, \cdots, \bm c_m) = \begin{bmatrix} \bm\rho \\ \bm\eta_v \end{bmatrix} + \bm v,
\end{equation}
where $\bm v$ represents the measurement noise with  covariance $\bm R=E[\bm v \bm v^T]$. The pose has to be resolved to minimize the distance between the two data sets through the  following least squares programming~\cite{Besl-Mckay-1992}
\begin{subequations} \label{eq:min_distance}
\begin{align}
\varepsilon = &\min_{\bm y} \sum_{i=1}^m \| \bm A(\bm\eta) \bm c_{i} +
\bm\rho - \bm d_{i} \|^2. \\
& \mbox{s.t.:} \; \bm\eta^T \bm\eta =1
\end{align}
\end{subequations}
Here, the variable $\varepsilon$ represents the ICP metric fit error, and $\bm A(\bm\eta)$ is the rotation matrix corresponding to quaternion $\bm\eta$, which can be computed in a similar manner to \eqref{eq:R}. It will be demonstrated later that the metric fit error $\varepsilon$ plays a critical role in fault detection and recovery of the vision system.
There are several algorithms available to solve the optimization problem \eqref{eq:min_distance}, such as the q-Method which computes the optimal quaternion as the eigenvector corresponding to the maximum eigenvalue \cite{Horn-1987}. Suppose the centroids of the points data sets are 
\begin{equation}
\bar{\bm c} =\frac{1}{m} \sum_{i=1}^m \bm c_i \qquad \mbox{and} \qquad  \bar{\bm d} =\frac{1}{m} \sum_{i=1}^m \bm d_i.
\end{equation}
Also define $4 \times 4$ matrix $\bm G$ with the following  construct
\begin{equation}
\bm G = \begin{bmatrix} \bm D + \bm D^T - \mbox{tr}(\bm D) \bm I  & \bm z \\ \bm z^T & \mbox{tr}(\bm D)  \end{bmatrix},
\end{equation}
where $\bm D=\sum_{i=1}^m (\bm c_i - \bar{\bm c}) (\bm d_i - \bar{\bm d})^T$ and $\bm z=\sum_{i=1}^m (\bm c_i - \bar{\bm c}) \times (\bm d_i - \bar{\bm d})$.
Then, it can be shown that the quaternion solution for the quadratic optimization programming  \eqref{eq:min_distance} is equal to the normalized eigenvector of $\bm G$ with the
largest eigenvalue, i.e., the solution of
\begin{equation} \label{eq:Glambda}
\bm G \bm\eta = \lambda_{\rm max} \bm\eta .
\end{equation}
Next, we can proceed with computation of the translation by
\begin{equation} \label{eq:rho=Ac}
\bm\rho=\bar{\bm d} - \bm A(\bm\eta) \bar{\bm c}
\end{equation}

From the kinematics and dynamics equations \eqref{eq:muOtimesq}, \eqref{eq:rho}, \eqref{eq:dot_q}, \eqref {eq:target_dyn}, and the registration equations \eqref{eq:Glambda} and \eqref{eq:rho=Ac}, the system's dynamic and nonlinear observation equations can be described in the following compact form.
\begin{subequations} \label{eq:observer_model}
\begin{equation}
\dot{\bm x} = \bm f(\bm x) + \bm L(\bm x) \bm w
\end{equation}
\begin{equation}
\bm y(\bm c_1 , \cdots, \bm c_m) = \bm h(\bm x) + \bm v
\end{equation}
Here, vector $\bm w^T= [\bm w_{\tau}^T \;\; \bm w_f^T]^T$ represents the overall process noise with covariance matrix $\bm W =E[\bm w \bm
w^T]=\mbox{diag}(\sigma_{\tau}^2 \bm I, \sigma_{f}^2 \bm I)$, and
\begin{align} \label{eq:f(x)}
\bm f(\bm x) &= \begin{bmatrix} \frac{1}{2} \mbox{vec} \big( \bm\Omega(\bm\omega) \bm q \big) \\ \bm\phi(\bm\omega, \bm\sigma)  \\
\dot{\bm\rho}_o \\
 \bm 0 \end{bmatrix}, \quad   \bm L(\bm x)  =\begin{bmatrix} \bm 0 & \bm 0 \\ \bm B(\bm\sigma) & \bm 0 \\ \bm 0 & \bm 0 \\ \bm 0 & \bm I \\ \bm 0 & \bm
 0\end{bmatrix}\\ \label{eq:h(x)}
\bm h(\bm x) &= \begin{bmatrix} \bm\rho_o + \bm A(\bm q) \bm\varrho \\ \mbox{vec} (\bm\mu \otimes \bm q)  \end{bmatrix}
\end{align}
\end{subequations}
Here, function $\mbox{vec}(\cdot)$ takes a quaternion and then returns its vector part.  Suppose $\hat{\bm q}$ represent the estimated quaternion and subsequently define small quaternion variable $\delta \bm q = \hat{\bm q}^{-1} \otimes \bm q$
to be  used as the states of linearized system 
and quaternion variation $\delta \bm\mu = \bm \mu \otimes \hat{\bm\mu}^{-1}$ is similarly defined. Then, we  can develop a constrained Kalman filter estimator to
estimate the unknown variables based on linearized model of \eqref{eq:observer_model} while respecting the constraints \eqref{eq:Dsigma<1}. Define
$\delta \hat{\bm x}_k^-$  and $\delta \hat{\bm x}_k^+$ as the {\em aprioir} and {\em aposteriori} estimates of the state vector at time $t_k$ \cite{Aghili-Salerno-2011}. Then, the estimation
update is given by
\begin{equation}
\bm e_k =\bm y_k - \bm h(\hat{\bm x}_k^-)
\end{equation}
\begin{equation} \label{eq:innovation}
\delta \hat{\bm x}_k^+ = \delta \hat{\bm x}_k^- + \mbox{th}(\varepsilon) \bm\Lambda_k \bm K^u_k \bm e_k
\end{equation}
Here, $\mbox{th}(\cdot)$ is a threshold function
\begin{equation} \label{eq:th}
\mbox{th}(\varepsilon) = \left\{ \begin{array}{ll} 1  \quad & \mbox{if} \quad  \varepsilon < \varepsilon^*\\
0 & \mbox{otherwise} \end{array} \right.
\end{equation}
whose output indicates whether the registration process is healthy or faulty,
and $\bm\Lambda_k=\mbox{diag}(1, 1, \cdots, \Lambda_{1_k}, \Lambda_{2_k}, \cdots, 1,1)$ where
\begin{equation} \label{eq:Lambda_k}
\Lambda_{i_k} =\left\{ \begin{array}{ll}
\mbox{sgn}(\bm k_{i_k}^T \bm e_k) -\frac{\hat{\sigma}_{i_k}^-}{\bm k_{i_k}^T \bm e} & \quad \mbox{if} \quad |\bm k_{i_k}^T \bm e_k| >1 \\
1 & \quad \mbox{otherwise} \end{array} \right.  \quad i=1,2
\end{equation}
and  $\bm k^T_{1_k}$ and $\bm k^T_{2_k}$ are the last two row vectors of the unconstrained gain matrix $\bm K_k^u$; see the Appendix for details.

\begin{figure}
\center{\includegraphics[width=12cm]{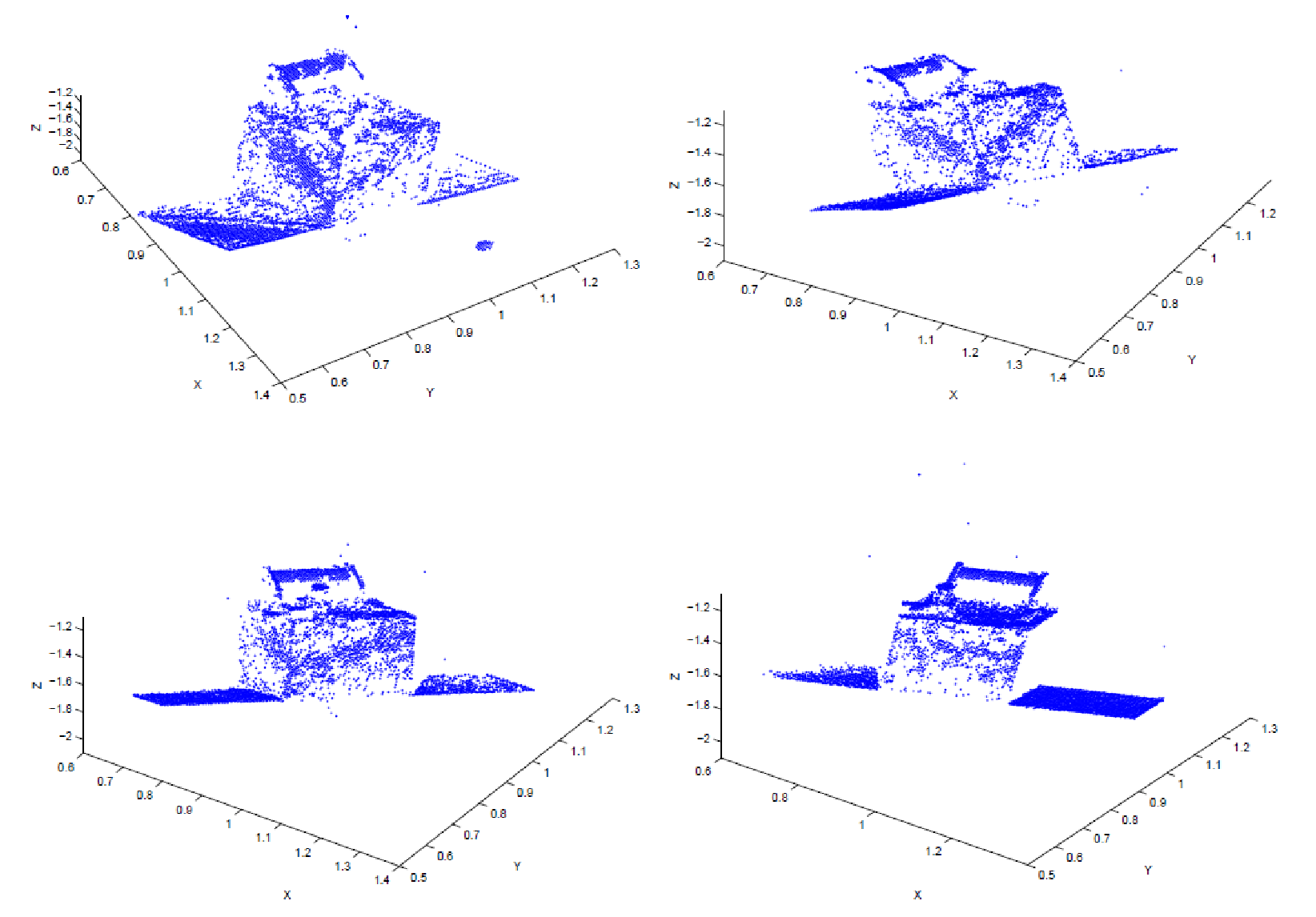}}  
\caption{Typical registered point-cloud data acquired by scanning the satellite mock-up at different poses.} \label{fig:scan}
\end{figure}

The propagation of the state vector is obtained from the nonlinear model
\begin{equation}
\hat{\bm x}_{k+1}^-  = \hat{\bm x}_{k}^+ +
\int_{t_k}^{t_{k}+t_{\Delta}} \bm f(\bm x)\,{\text d} t
\end{equation}
Equation \eqref{eq:th} constitutes a simple fault-detection logic based on comparing the matching error $\varepsilon$ against the threshold $\varepsilon^*$. Clearly, whenever vision registration fault is detected, then the state update process in not affected by the observation information, i.e.,
\begin{equation}
\varepsilon > \varepsilon^* \; \Longrightarrow \;  \hat{\bm x}_k^+ = {\bm x}_k^- \quad \wedge \quad  \hat{\bm x}_{k+1}^-  = \hat{\bm x}_{k}^- +
\int_{t_k}^{t_{k}+t_{\Delta}} \bm f(\bm x)\,{\text d} t.
\end{equation}
In other words, the estimator relies on the dynamics model for pose estimation until ICP becomes convergent for estimation update. As will be later discussed  in the experiment Section~\ref{sec:post-experiments}, four typical sets of point-cloud data registered by the vision sensor at different poses are illustrated in Fig.~\ref{fig:scan}. It is apparent from the figure that the quality of the acquired 3D images, e.g., the number or returned points and outliers, varies from one scan to another.

\section{Pre-Capturing Trajectory Planning} \label{sec:pre-capturing}
This section presents the development of an optimal robot guidance method for rendezvous and smooth interception of tumbling/moving objects based on visual feedback. It is assumed that the Attitude and Orbit Control System (AOCS) of the servicer compensates for the dynamic coupling between the motion of its robot arm and base, ensuring that trajectory planning is not affected. It is worth noting that the trajectory planning of the robot during pre- and post-capture phases is executed in the task space. As a result, appropriate inverse-kinematic techniques should be implemented to address any complications that may arise due to singularities and joint limits~\cite{Han-Park-2013,Aghili-2005}.
The position of the end-effector and the capture point are represented by $\bm r$ and $\bm\rho$, respectively--refer to Fig.~\ref{fig:robot_target}.a. To prevent impact at the end of the capture phase, it is imperative that the robot's end-effector intercepts the target's grapple point with zero relative velocity.
Suppose the optimal trajectory is manifested  by
\begin{equation} \label{eq:sys_xr}
\ddot{\bm r} =\bm u_1,
\end{equation}
which can be formally rewritten as $\dot{\bm x}_1 =[ \dot{\bm r}^T  \; \bm u_1^T ]^T$ where  $\bm x_1^T=[\bm r^T \; \dot{\bm r}^T]$. Denting terminal time $t_1$, one can write the terminal condition as $\bm\psi(t_1)= \bm 0$, where
\begin{equation} \notag
\bm\psi(t) = \begin{bmatrix} \bm r(t) - \bm\rho(t) \\  \dot{\bm r}(t) - \dot{\bm\rho}(t) \end{bmatrix}
\end{equation}
The terminal position and velocity can be calculated by integration of the acceleration
\begin{equation} \notag
\ddot{\bm\rho} =    \bm A({\bm q}) \big( {\bm\omega} \times( {\bm\omega} \times \hat{\bm\varrho}) + \bm\phi({\bm\omega}, \hat{\bm\sigma})\times \hat{\bm\varrho} \big)
\end{equation}
given initial conditions $\dot{\bm\rho}(t_0) = {\dot{\bm\rho}}_{o} + \bm A({\bm q}) \big( {\bm\omega} \times \hat{\bm\varrho} \big)$ and ${\bm\rho}(t_0) = {\bm\rho}_{o}
+ \bm A({\bm q}_k) \hat{\bm\varrho}$, where $\hat{\bm\sigma}$ and $\hat{\bm\varrho}$ denote the estimated values of the corresponding variables.  Another constraint is that the target's capturing fixture should  be accessible by the robotic hand for capturing at the time of
capture. In other words, the target satellite must be with right orientation at the time of interception for LOS obstruction avoidance of the grasping point on the target. In order to enforce the accessibility constraint, we define
angle $\alpha$ made between the normal vector $\bm k$ on the surface of capturing fixture and the camera line of sight $\bm\rho$. At the time of grasping $t_1$, when $\bm\rho(t_1)=\bm r (t_1)$, $\alpha$ becomes the angle between the normal vector and  the end-effector position vector $\bm r$, see Fig.~\ref{fig:robot_target}. Then, one can conclude that best alignment of the target satellite for capturing accessibility is tantamount to minimize the following function
\begin{equation} \notag
\varphi(t) = -w \cos \alpha(t) = - w\frac{\bm\rho^T}{\| \bm\rho \|}  \bm A(\bm q) \bm k
\end{equation}
where $w$ is a weight. 

In the following analysis, we seek a time-optimal solution to the input $\bm u_1$ subject to the acceleration limit $\|\ddot{\bm r} \|\leq a_{1 \rm max}$ and the aforementioned terminal constraints, i.e.,
\begin{subequations} \label{eq:optimal_formulation}
\begin{align} \label{eq:min_time}
\mbox{minimize}   &\qquad \varphi( t_1) +\int_{t_0}^{t_1} 1 \; d t \\ \label{eq:a_max}
\mbox{subject to:}   &  \qquad \| \bm u_1(\tau) \| \leq a_{1\rm max}  \qquad t_0 \leq t \leq t_1 \\ \label{eq:terminal1}
& \qquad \bm\psi_1(t_1) = \bm 0
\end{align}
\end{subequations}
It's worth mentioning that our visual servoing setup for the positioning of the
camera is eye-to-hand, which means the camera is placed at a fixed point in the workspace. Therefore, the vision system is not affected by the the robot velocity. However, for the case of eye-in-hand setup where the camera is  installed on the robot end-effector, the velocity constraint might be included in the optimal control formulation \eqref{eq:optimal_formulation} to avoid failure of the vision system.

Defining the vector of Lagrangian multiplier as $\bm\lambda_1$, one can write the expression of the system Hamiltonian in the pre-capturing phase as follows:
\begin{equation} \label{eq:H1}
 H_1 = 1 + \bm\lambda_1^T \dot{\bm x}_1
\end{equation}
Note that the unity in the expression of the right-half-side of \eqref{eq:H1} arises from $g=1$ in the cost function \eqref{eq:min_time}. The optimal control
theory~\cite{Anderson-Moore-1990} dictates that the time-derivative of the costate  must satisfy
\begin{equation} \label{eq:lamb1}
\dot{\bm\lambda}_1 = -\frac{\partial H_1}{\partial \bm x_1} \quad \mbox{hence} \quad \bm\lambda_1^* = \begin{bmatrix} \bm a_1 \\ - \bm a_1 \tau + \bm a_2 \end{bmatrix},
\end{equation}
where $^*$ indicates optimal values, the $6 \times 1$ vector  $\bm a^T=[\bm a_1^T , \; \bm a_2^T]$ contains the constants to be found later from the boundary conditions. Thus, by virtue of \eqref{eq:H1} and \eqref{eq:lamb1}, we can say
\begin{equation} \label{eq:H*}
H_1(\bm x_1^*, \bm\lambda_1^*, \bm u_1) = 1 + \bm a_1^T \dot{\bm r} + (-\bm a_1^T \tau + \bm a_2^T ) \bm u_1.
\end{equation}
The Pontryagin's principle dictates that the optimal input $\bm u_1^*$ satisfies
\begin{equation} \notag
\min_{\bm u_1} H_1(\bm x_1^*, \bm\lambda_1^*, \bm u_1).
\end{equation}
Therefore, in view of the acceleration limit constraint \eqref{eq:a_max} and expression \eqref{eq:H*}, the optimal control input in the pre-capturing phase must take the following structure
\begin{equation} \label{eq:u}
\bm u_1^*  =-\frac{-\bm a_1 \tau  + \bm a_2}{\| -\bm a_1 \tau  + \bm a_2 \|} a_{1 \rm max} \quad t_0 \leq t \leq t_1
\end{equation}
The optimal terminal time $t_1$ along with constant vectors $\bm a_1$ and $\bm a_2$ remain to be found. The transversality condition dictates the following identity
\begin{equation} \label{eq:H1=0}
\frac{\partial \varphi }{\partial t_1} + H^*_1(t_1) =0
\end{equation}
where
\begin{equation} \notag
 H^*_1(t_1) = 1+ \bm a_1^T \dot{\bm r}(t_1) + \|  \bm a_1 t_1 - \bm a_2 \| a_{1\rm max}
\end{equation}
\begin{equation} \label{eq:partial_varphi}
\frac{\partial \varphi }{\partial t_1} = \left( \frac{\partial \varphi^T}{\partial \bm\xi} \dot{\bm\xi} \right)_{\!\!t_1}.
\end{equation}
Here,  vector $\bm\xi^T=[\bm r^T \;\; \bm q^T]$ contains the position and orientation, and the vectors in the right-hand side of \eqref{eq:partial_varphi} are given by
\begin{align*}
\left( \frac{\partial \varphi}{\partial \bm\xi} \right)_{\! t_1} & = \frac{2w}{\| \bm r (t_1) \|} \begin{bmatrix} \frac{\bm r \times(\bm r \times \bm A \bm k)}{2\| \bm r \|^2} \\
 q_o \bm k \times \bm r +  (\bm q_v \times \bm k) \times  \bm r + (\bm q_v \times \bm r) \times \bm k \\
\bm r^T(\bm q_v \times \bm k + 2 q_o \bm k) \end{bmatrix}_{t_1} \\
\left( \dot{\bm\xi} \right)_{\! t_1} &= \begin{bmatrix} \dot{\bm r} \\ \frac{1}{2} \bm\Omega(\bm\omega) \bm q \end{bmatrix}_{t_1},
\end{align*}
where $\bm r(t_1)=\bm\rho(t_1)$ and  $\dot{\bm r}(t_1)=\dot{\bm\rho}(t_1)$. Finally applying the terminal conditions \eqref{eq:terminal1} to \eqref{eq:u} and combining the resultant equations with \eqref{eq:H1=0}, we arrive at the following error
equation in terms of seven unknowns $\{\bm a, \; t_1 \}$, i.e.,
\begin{equation}  \notag 
\| \bm e_1(\bm a ,t_1) \| =0, \quad \bm e_1(\bm a, t_1) =  \begin{bmatrix} \bm r(\bm a, t_1)- \bm\rho(t_1) \\
\dot{\bm r}(\bm a, t_1)- \dot{\bm\rho}(t_1) \\
 \left( \frac{\partial \varphi^T}{\partial \bm\xi} \dot{\bm\xi} \right)_{\!\!t_1} + H_1(\bm a, t_1) \end{bmatrix}.
\end{equation}
The above equations can be solved for unknowns  $\{\bm a, \; t_1 \}$  by utilizing  a numerical technique, e.g., the Newton-Raphson method.

\section{Post-Capturing Trajectory Planning} \label{sec:post-capturing}
Fig.~\ref{fig:robot_target}.b schematically illustrates the post-capturing  operation, which starts after completion of the capturing phase. In this section, we seek  another optimal trajectory planning for the post-capturing phase. The control objective is to damp out the momentums of the tumbling and drifting target as quickly as possible without applying excessive force and torque.
Suppose the target linear velocity, $\bm v$, angular velocity, $\bm\omega$, as well as the exerted force, $\bm f_e$, and torque, $\bm\tau_e$, are all expressed in the body coordinate frame attached to the target at its location of CoM. Then, the equations of the motion of the target in the post-capturing phase is described by
\begin{subequations} \label{eq:dot_ph}
\begin{align} \label{eq:dotp}
\dot{\bm\upsilon}  & =  - \bm\omega \times \bm\upsilon  + \frac{1}{m} \bm f_e \\ \label{eq:Euler}
\dot{\bm\omega}  & =  \bm\phi(\bm\omega,\bm\sigma)   + \frac{1}{\mbox{tr}(\bm I_c)} \bm B(\bm\sigma)( \bm \tau_e - \bm\varrho \times \bm f_e ).
\end{align}
\end{subequations}
Denoting the system states in the post-capture phase by vector $\bm x_2^T=[\bm\upsilon^T \; \bm\omega^T]$ and the control input $\bm u_2^T = [
\bm f_e^T \;\;  \bm\tau_e^T]$, we are interested in
optimal input trajectories  $\bm u_2^*$ which damp out the target's linear and angular velocities  at the time of interception, i.e.,  $\bm\upsilon(t_1)$ and
$\bm\omega(t_1)$, in minimum time subject to maximum magnitude limits of the input force and torque to be  $f_{{\rm max}}$ and $\tau_{{\rm max}}$, respectively. Note that initial linear and angular velocities of the target at the time of interception, i.e.,  $\bm v(t_1)$ and $\bm\omega(t_1)$, are equal to those the robot end-effector on the servicer and therefore they can be calculated from the robot joint rates. Thus
\begin{subequations}
\begin{align} \notag
\mbox{minimize} & \qquad \int_{t_1}^{t_2}  \;  dt\\ \label{eq:f_max}
\mbox{subject to:} & \qquad \| \bm f_e \| \leq f_{{\rm max}} \\ \label{eq:tau_max}
& \qquad \| \bm\tau_e \| \leq \tau_{{\rm max}} \\ \label{eq:terminal}
& \qquad \bm\psi_2(t_2) = 0
\end{align}
\end{subequations}
where $\bm\psi_2(t)=\bm x_2(t)$ is the final condition of the post-capturing phase. The Hamiltonian of the
system in post-capturing phase can be written as
\begin{align}  \label{eq:Hamilton}
H_2 &= 1 + \bm\lambda_2^T \dot{\bm x}_2 \\  \notag
 &=  1 - \bm\lambda_2'^T (\bm\omega \times \bm\upsilon) +  \bm\lambda_2''^T \bm\phi(\bm\omega) + \frac{1}{\mbox{tr}(\bm I_c)} \bm\lambda_2''^T \bm B \bm\tau_e \\ \notag
 & + \Big(\frac{1}{m}\bm\lambda_2' - \frac{1}{\mbox{tr}(\bm I_c)} \bm\varrho \times \bm B \bm\lambda_2'' \Big)^T \bm f_e
\end{align}
Then, the time-derivative of the corresponding costates is dictated by the following partial derivative equation
\begin{equation} \notag
\dot{\bm\lambda}_2  = -  \frac{\partial H_2}{\partial \bm x_2},
\end{equation}
and thus  we have
\begin{equation}\label{eq:dot_lambda}
\dot{\bm\lambda}_2  = \begin{bmatrix} -[\bm\omega \times] & \bm 0 \\ [\bm\upsilon \times] & \frac{1}{\mbox{tr}(\bm I_c)} \frac{\partial \bm\phi^T}{\partial \bm\omega}
\end{bmatrix} \bm\lambda_2,
\end{equation}
where
\begin{equation}
\frac{\partial \bm\phi}{\partial \bm\omega} =\begin{bmatrix} 0 & \sigma_1 \omega_z & \sigma_1 \omega_y \\ \sigma_2 \omega_z & 0 & \sigma_2 \omega_x \\ -\frac{\sigma_1 +
\sigma_2}{1+ \sigma_1 \sigma_2} \omega_y & -\frac{\sigma_1 + \sigma_2}{1+ \sigma_1 \sigma_2} \omega_x & 0 \end{bmatrix}.
\end{equation}
Moreover, allowable trajectories of the optimal control input should minimize the Hamiltonian function,
according to the {\em Pontryagin's Minimum Principle} of the optimal control theory. That is
\begin{equation} \label{eq:Pontragin}
\min_{\bm u_2} \; H_2(\bm x_2^*, \bm\lambda_2^*, \bm u_2)
\end{equation}
subject to inequality and equality constraints \eqref{eq:f_max}, \eqref{eq:tau_max}, and \eqref{eq:terminal}. The expression of the Hamiltonian \eqref{eq:Hamilton} can be  concisely written  by
\begin{align}  \label{eq:Hamilton2}
H_2 &= c  + \frac{1}{\mbox{tr}(\bm I_c)} \big[ \bm p_1^T \bm\tau_e + \bm p_2^T \bm f_e \big]
\end{align}
where the auxiliary variables are defined by $c=1 - \bm\lambda_2'^T (\bm\omega \times \bm\upsilon) +  \bm\lambda_2''^T \bm\phi(\bm\omega)$, $\bm p_1=\bm B  \bm\lambda_2''$, $\bm p_2 = \kappa^2 \bm\lambda_2 - \bm\varrho \times \bm B \bm\lambda_2''$, and
\begin{equation} \notag
\kappa = \sqrt{\frac{\mbox{tr}(\bm I_c)}{m}}
\end{equation}
is the Euclidean norm of gyradius of the satellite body about all three axes, i.e., $\kappa=\sqrt{\kappa_x^2+\kappa_y^2+\kappa_z^2}$.
Clearly the  expression of the system Hamiltonian in \eqref{eq:Hamilton2}  is minimized when the direction of the torque and force vectors are aligned in opposite direction of the axillary vectors $\bm p_1$ and $\bm p_2$, respectively. That is the optimal  torque and force should be proportional to the unit vectors  $- \bm p_1/ \| \bm p_1 \|$ and $- \bm p_2/ \| \bm p_2 \|$, respectively. Moreover, since the maximum magnitude that vectors $\bm \tau_e$ and $\bm f_e$ can take are $\tau_{\rm max}$ amd $f_{\rm max}$, one can infer that the optimal force and torque control inputs in the post-capturing phase must have the following constructs in order to minimize the Hamiltonian
\begin{subequations} \label{eq:opt_inputs}
\begin{align}
\bm\tau_e^* & = - \frac{\bm p_1}{\| \bm p_1 \|} \tau_{{\rm max}}= -\frac{\bm B \bm\lambda_2''}{\| \bm B \bm\lambda_2'' \|} \tau_{{\rm max}}, \\
\bm f_e^* &= - \frac{\bm p_2}{\| \bm p_2 \|} f_{{\rm max}}= -\frac{\kappa^2 \bm\lambda_2' + \bm\varrho \times \bm B \bm\lambda_2''}{\| \kappa^{2} \bm\lambda_2' + \bm\varrho \times  \bm B \bm\lambda_2'' \| } f_{{\rm
max}}
\end{align}
\end{subequations}
Then, upon substitution of \eqref{eq:opt_inputs} into  \eqref{eq:dot_ph}, we arrive at the optimal motion trajectories for the post-capturing maneuvering
\begin{subequations} \label{eq:optimal_ph}
\begin{align} 
\dot{\bm\upsilon}^*  & = - \bm\omega^* \times \bm\upsilon^* -\frac{\kappa^{2} \bm\lambda_2' +  \hat{\bm\varrho} \times  \bm B \bm\lambda_2'' }{\| \kappa^{2} \bm\lambda_2' +
\hat{\bm\varrho} \times \bm B \bm\lambda_2'' \| } a_{2\rm max} \\
\dot{\bm\omega}^*  & = \bm\phi(\bm\omega^*, \hat{\bm\sigma}) +  \frac{\bm B^{2} \bm\lambda_2''}{\| \bm B \bm\lambda_2'' \| }  \gamma_{\rm max} \\ \notag
&- \bm B \frac{\hat{\bm\varrho} \times \bm\lambda_2' + \kappa^{-2} \hat{\bm\varrho} \times (\hat{\bm\varrho} \times \bm B \bm\lambda_2'') }{\|\kappa^{2} \bm\lambda_2' +  \hat{\bm\varrho} \times
\bm B \bm\lambda_2'' \| } a_{2\rm max} , \\
\dot{\bm q}^* &= \frac{1}{2} \bm\Omega(\bm\omega^*) \bm q^*\\
\bm u_2^* & = \bm A(\bm q^*) \big( \dot{\bm\upsilon}^* + \dot{\bm\omega}^* \times \hat{\bm\varrho} + \bm\omega^* \times(\bm\omega^* \times \hat{\bm\varrho}) \big)
\end{align}
\end{subequations}
where $\hat{\bm\sigma}$ and $\hat{\bm\varrho}$ are the estimated values  of the corresponding unknown parameters, while
\begin{equation} \label{eq:user-defined}
a_{2\rm max} := \frac{f_{{\rm max}}}{m} \quad \mbox{and} \quad  \gamma_{\rm max} :=  \frac{\tau_{{\rm max}}}{\mbox{tr}(\bm I_c)}
\end{equation}
are the user-defined parameters corresponding to the maximum linear and angular accelerations in the post-capturing phase. It is worth noting that combining the maximum force and torque parameters with the mass and trace of moment of inertia tensor parameters effectively eliminates the requirement for precise knowledge of the unidentifiable parameters that cannot be directly determined from observing the target's motion. If the target's mass and trace of moment of inertia tensor are not precisely known, conservative upper-bound values can be used to determine the user-defined parameters for maximum acceleration \eqref{eq:user-defined}. By incorporating these acceleration parameters into the optimization, the maximum force and moment values can be constrained to not exceed their limits, even if the estimation of the target's mass and trace of moment of inertia tensor are imprecise. However, this approach may result in a suboptimal time solution.

The differential equations \eqref{eq:optimal_ph}  together with \eqref{eq:dot_lambda}
can be solved upon knowing the initial value of the costate vector. Since the optimal control system is with open-end time, the transversality condition implies that the Hamiltonian calculated  over time interval  $t_1 \leq  t \leq t_2$ must be zero. That is to say
\begin{equation} \label{eq:H2=0}
H_2(t_2)=0.
\end{equation}
Now, we can obtain additional equation in order to calculate the optimal terminal time upon substitution of \eqref{eq:opt_inputs}  into \eqref{eq:Hamilton} and using identity \eqref{eq:H2=0}. That is
\begin{align}\label{eq:H=0}
H_2(t) = & 1-(\bm\omega \times \bm\upsilon)^T \bm\lambda_2'  + \bm\phi^T(\bm\omega)^T \bm\lambda_2'' \\ \notag
& - \| \bm B \bm\lambda_2'' \| \tau_{{\rm max}} - \| \kappa^{-2}\bm\lambda_2' + \bm\varrho \times \bm B \bm\lambda_2'' \| f_{{\rm max}} = 0
\end{align}
The above set of equations can be numerically solved to obtain the initial value of the costate  and the final  time, i.e.,   $\{\bm\lambda_2(t_1), \; t_2\}$. The shooting method can be utilized to solve
this two-point boundary value problem (TPBVP) through zeroing the terminal error calculated by
numerical integration of \eqref{eq:dot_lambda} and \eqref{eq:optimal_ph}. To this effect, the  error function is defined as
\begin{equation} \label{eq:error_function}
\| \bm e_2 (t_2) \| = 0 \;\;  \mbox{where} \;\; \bm e_2(t_2) = \begin{bmatrix} \bm\upsilon(\bm\lambda_2(t_1),t_2)\\ \bm\omega(\bm\lambda_2(t_1),t_2)\\ H_2^*(\bm\lambda_2(t_1),t_2) \end{bmatrix}.
\end{equation}
The above error function will vanish if the unknown variables $\{\bm\lambda_2(t_1), \; t_2\}$ take their correct values. To this end, a quasi-Newton method \cite{Boyarko-Yakimenko-2011} can be employed to find a numerical solution, e.g., by using Matlab function \verb"fminunc".

Fig.~\ref{fig:control2} illustrates the integration of two optimal control strategies that are associated with the pre- and post-capturing phases. This integration enables a smooth transition and facilitates real-time switching to the appropriate control system. The switching control system ensures a seamless transition from pre-capture control to post-capture control, triggered by a switching signal at the terminal time epoch $t_1$, and initialized with a snapshot of the system's states. Moreover, both controllers are continuously adjusted to the greatest extent possible by utilizing feedback obtained through vision data processing.

\section{Experiments} \label{sec:post-experiments}


The experimental setup of the satellite simulator described in \cite{Aghili-2023} is used to demonstrate the proposed robot guidance and control scheme. The scheme aims to capture and stabilize a satellite mockup that exhibits both translational and tumbling motions, while achieving the functional requirements in a simulated space environment through end-to-end robotic operation, including learning, pre-capturing, and post-capturing steps. 
This completes our previous work in \cite{Aghili-2011k}, which lacked experimental validation related to the post-capturing phase. To simulate the dynamic motion of a free-floating target satellite and a servicing robot \cite{Aghili-Namvar-2008,Aghili-Namvar-Vukovich-2006}, two manipulator arms are employed, as shown in Fig.~\ref{fig:lcs_cart}. In this experiment, Neptec laser scanner ~\cite{Samson-English-Deslauriers-Christie-2004} is placed at a fixed point in the workspace to generate 3D image data with the update rate of $2$~Hz for the eye-to-hand visual servoing method, see also Fig.\ref{fig:scan}.

\begin{figure}
\centering{\includegraphics[clip,width=8cm]{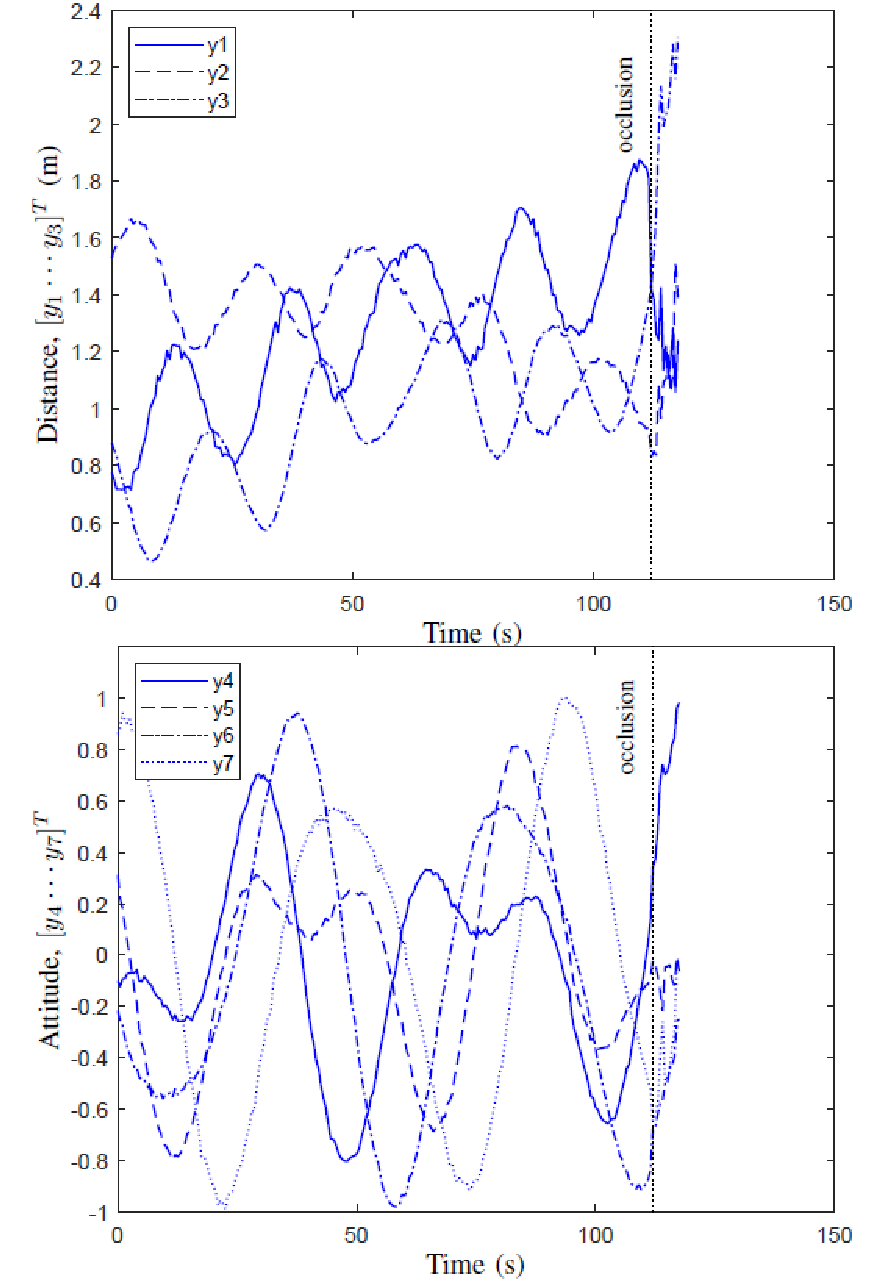}}
\caption{Target pose calculated by the image registration algorithm.}\label{fig:pose}
\end{figure}

\begin{figure}[h]
\centering{\includegraphics[clip,width=8cm]{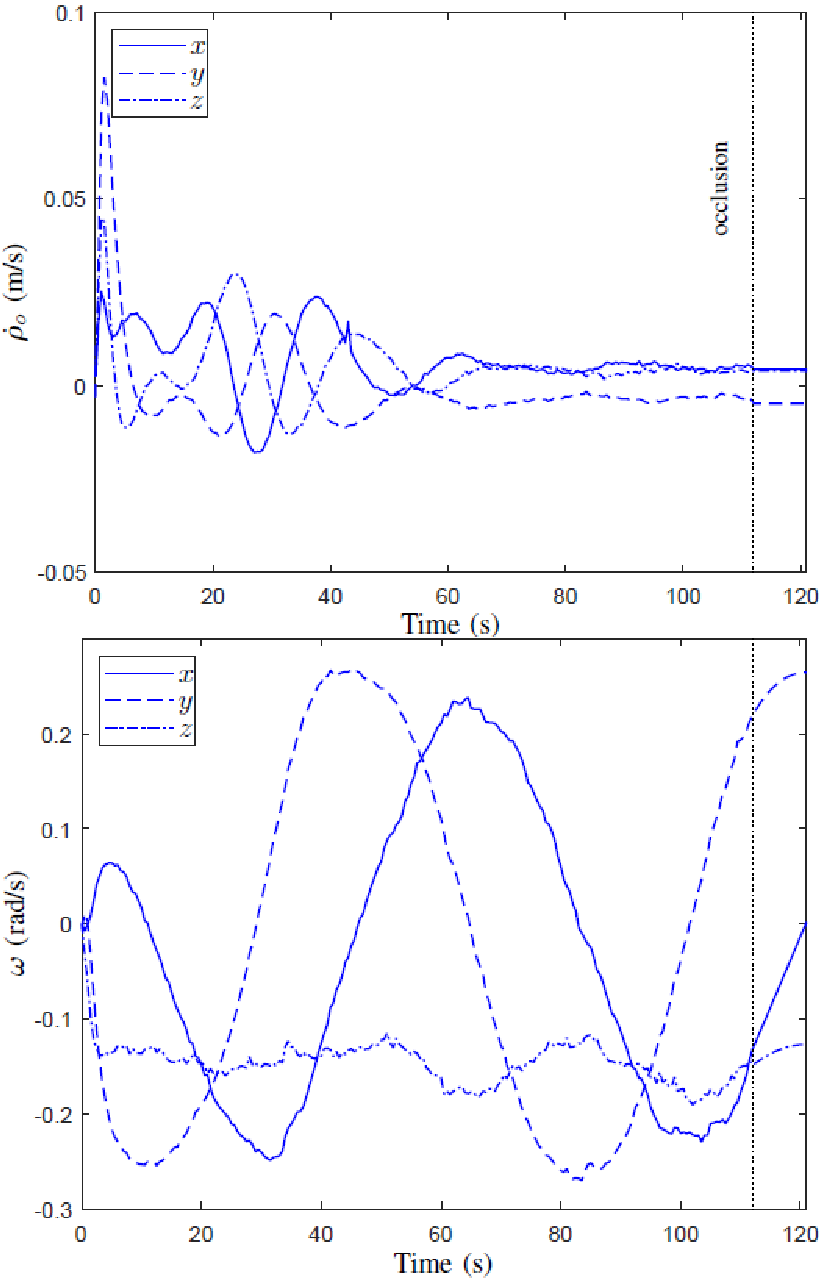}}
\caption{Estimation of the target's linear and angular velocities.} \label{fig:velocity}
\end{figure}

\begin{figure}
\centering{\includegraphics[clip,width=8cm]{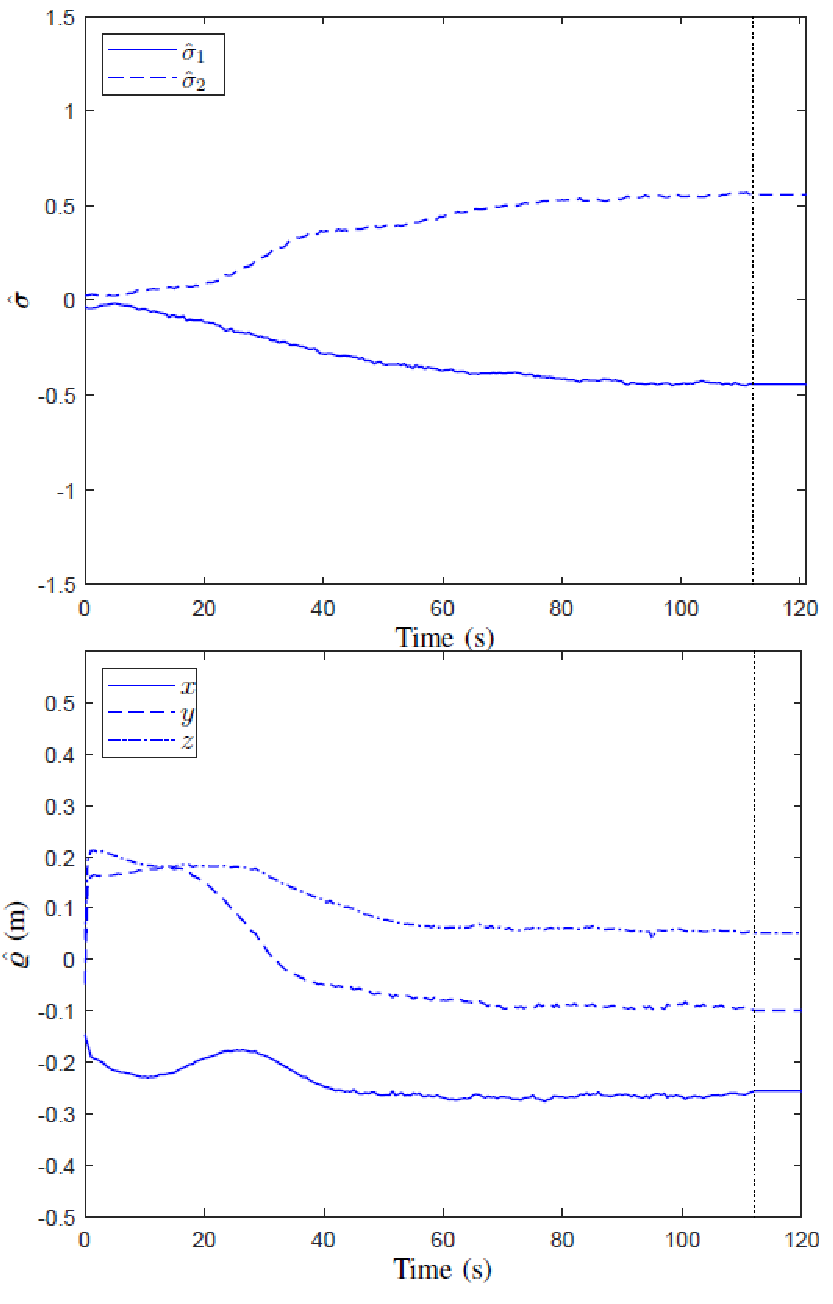}}
\caption{Time-histories of the estimated parameters against their actual values.}
\label{fig:param}
\end{figure}

\begin{table}
\caption{Simulated target parameters.}
\begin{center}
\begin{tabular}{cc}
\hline \hline
parameter & value\\
\hline
$m$ (kg) & 1600\\
$\bm{\varrho}$ (m) &   $[-0.25 \;\; -0.1 \;\; 0.05 ]^T$ \\
$\bm I_c$ (kg-m$^2$) & \mbox{diag}(400, \; 500; 700) \\
$\bm\sigma$ & $[-0.5 \; 0.6]^T$\\
\hline \hline
\end{tabular}
\end{center}
\label{tab:inertia_target}
\end{table}

One of the simulating manipulators has an end-effector that is mechanically connected to the satellite mockup, and it simulates the representative motion trajectories of the target satellite based on the inertial parameters listed in Table~\ref{tab:inertia_target}. The maximum contact force and torque are set to $f_{\rm max}=7.0$~N and $\tau_{\rm max}=8.0$~Nm, respectively. It should be noted that all identifiable inertial parameters of the target are unknown and are therefore estimated during the learning phase. We assume upper-bound values for the mass and trace of the moment of inertia tensor to be 1700~kg and 1800~Nm$^2$, respectively, which are approximately 10\% higher than the actual values specified in Table~\ref{tab:inertia_target}. Consequently, the maximum linear and angular accelerations in the post-capturing phase are set to $0.0035$~m/s$^2$ and $0.0045$~rad/s$^2$. The user-defined parameters of the optimal control for pre- and post-capturing manoeuvres are provided in Table\ref{tab:optimal_param}.

\begin{table}
\caption{User-defined parameters of the optimal control.}
\begin{center}
\begin{tabular}{cccc}
\hline \hline
Parameter & $a_{1\rm max}$ & $a_{2\rm max}$ &  $\gamma_{\rm max}$ \\
& (m/s$^2$)  & (m/s$^2$) & (rad/s$^2$) \\
\hline
Value & 0.01 & 0.0035 & 0.0045\\
\hline \hline
\end{tabular}
\end{center}
\label{tab:optimal_param}
\end{table}

\begin{figure}
\centering{\includegraphics[clip,width=8cm]{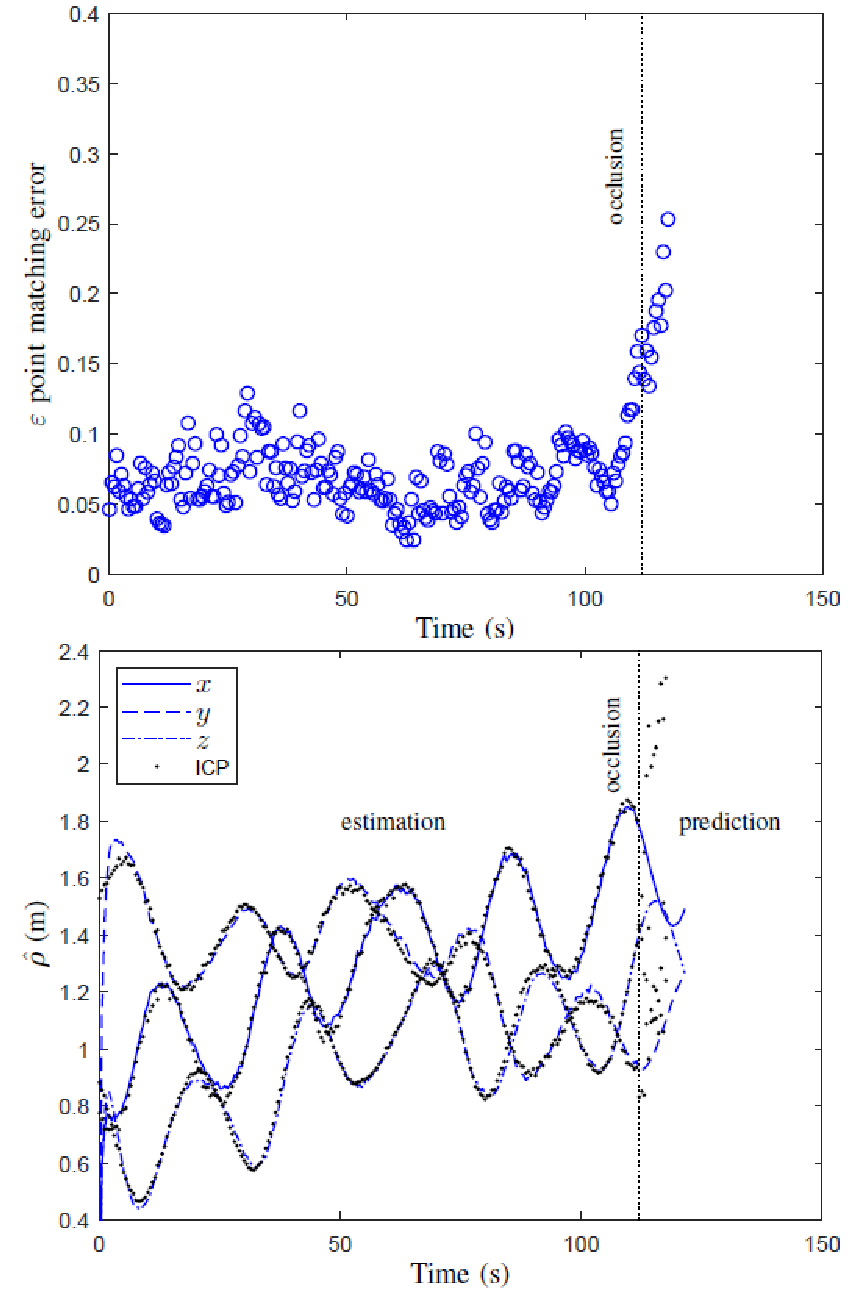}}
\caption{The ICP metric fit error (top) and predicted trajectory of the  grapple-fixture
position (bottom).} \label{fig:icp_eps}
\end{figure}

\begin{figure}
\centering{\includegraphics[clip,width=8cm]{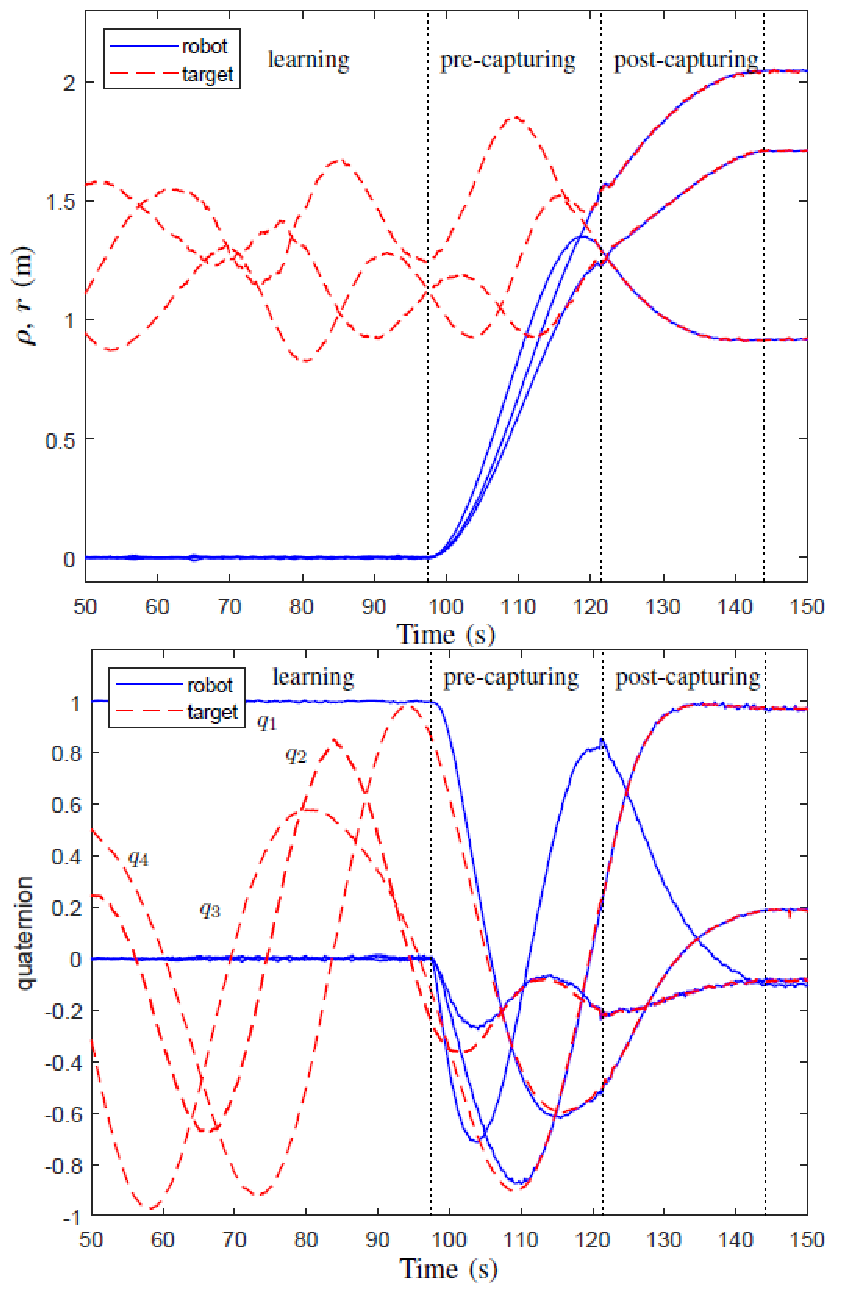}} \caption{The position and orientation
trajectories of the robot and the target.}\label{fig:capture_position}
\end{figure}

\begin{figure}
\centering{\includegraphics[clip,width=8cm]{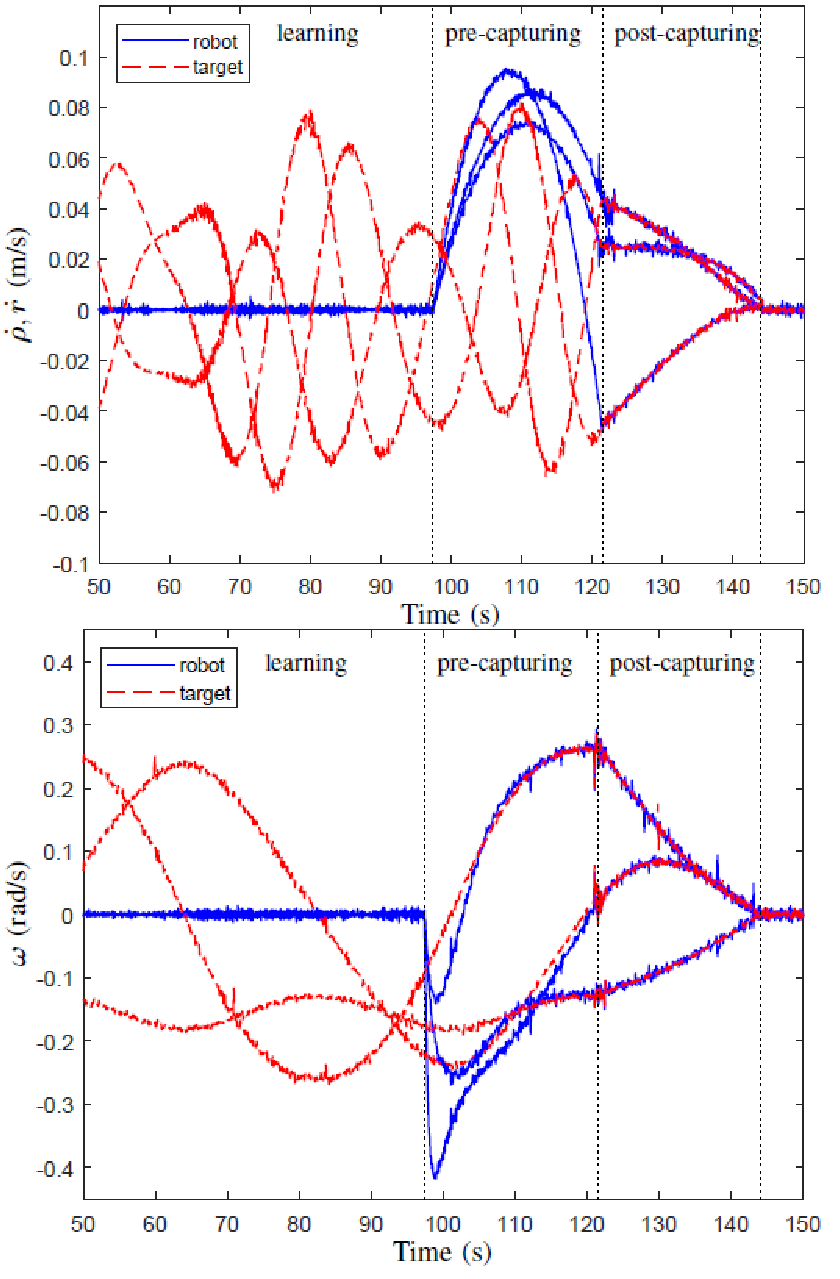}} \caption{Velocity trajectories of the robot and
the target.}\label{fig:capture_velocity}
\end{figure}

\begin{figure}
\centering{\includegraphics[clip,width=8cm]{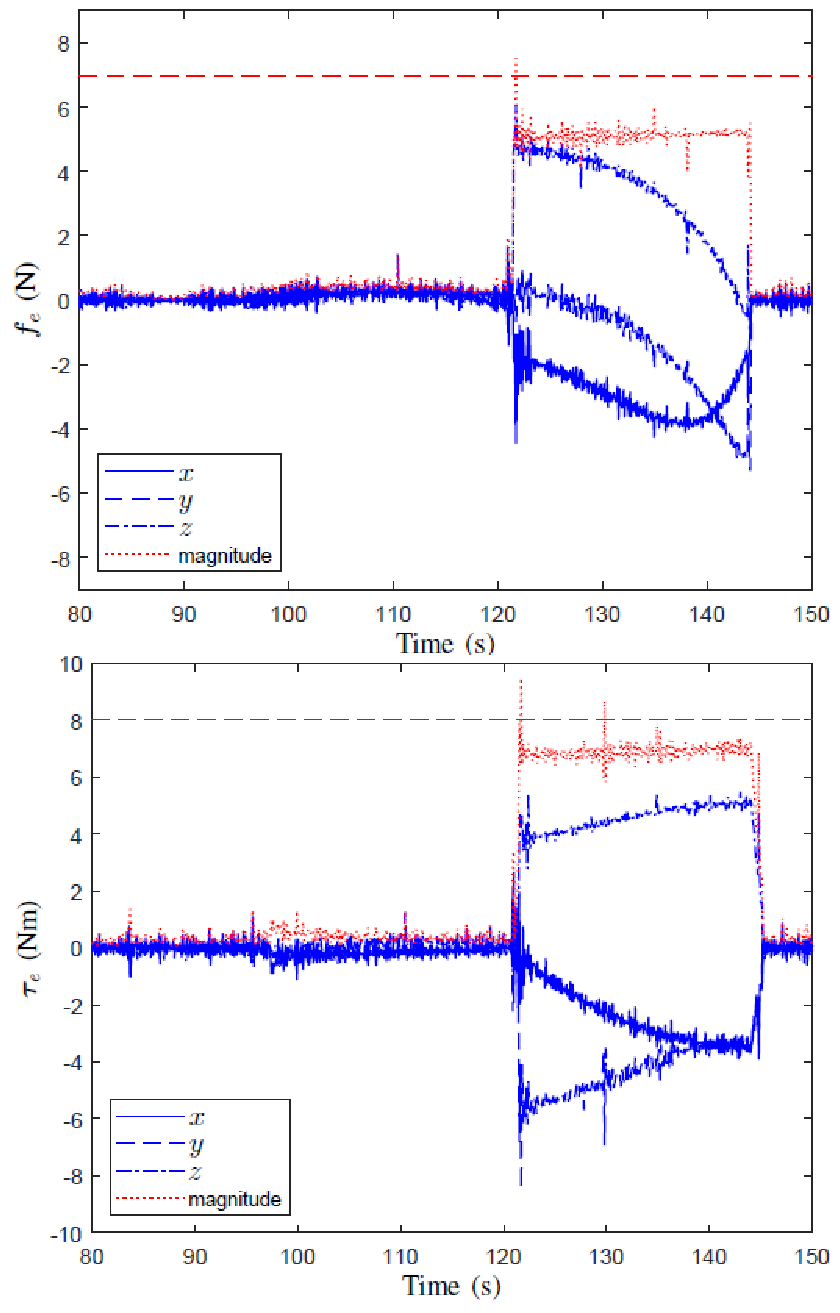}} \caption{Trajectories of the interaction force and moment during the post-capturing phase.}\label{fig:capture_force}
\end{figure}

\begin{figure}
\centering{\includegraphics[clip,width=8cm]{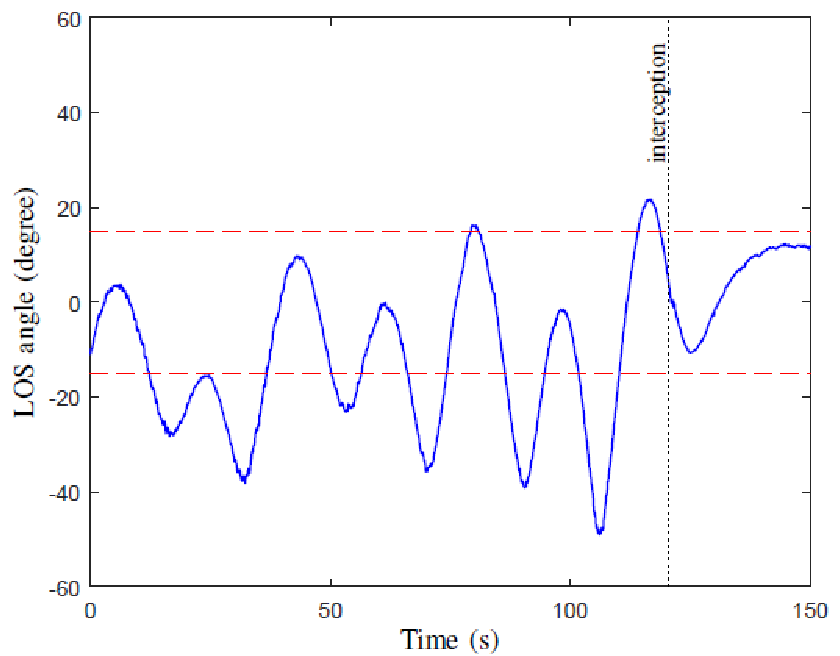}} \caption{Trajectory of the LOS angle.}\label{fig:LOS}
\end{figure}

\begin{table}
\caption{Timing and sequence of events.}
\begin{center}
\begin{tabular}{cccccc}
\hline \hline
Event   & convergence  & approach & occlusion & interception & stabilization \\
 & $T_c$ & $T_o$ & $T_{\rm oc}$ & $T_1$  & $T_2$ \\
\hline
Time  & 92.4~s & 97.4~s & 111.9 & 121.4~s & 144.3~s \\
\hline \hline
\end{tabular}
\end{center}
\label{tab:events}
\end{table}
The table listing the timing and sequence of events during the execution of the optimal guidance and control of the servicing robot can be found in Table~\ref{tab:events}. Additionally, Fig.~\ref{fig:pose} shows the pose trajectories of the target obtained from the 3D point-matching registration algorithm before and after the pre-capturing phase. The target pose, consisting of position and orientation, is calculated by the image processing using equation \eqref{eq:y}, where $y_1 \cdots y_7$ represent the individual elements of the pose.
It is important to note that the vision system fails before the completion of the pre-capturing maneuver at time $t=111.9$~sec. The failure of the vision system is attributed to the servicing robot's hand coming into the vision sensor's field-of-view, which inevitably causes the point-matching error. Nevertheless, prior to the trajectory planning and execution, the fault-tolerant estimator receives potentially erroneous data from the 3D vision registration algorithm and subsequently provides the best estimate of the target states, including linear and angular velocities, as well as its inertial parameters. These variables are incorporated in the pre- and post-capturing trajectory planning, and thus their accurate estimation is vital for the successful implementation of the overall robot guidance and control. The estimator's convergence during the learning phase is determined by continuously monitoring the Euclidean norm of the covariance matrix. When the norm reaches a sufficiently small value, the estimator is considered converged. In this experiment, the estimator converged at $t=92.4$~sec. Therefore, the initial time for the pre-capturing manoeuvre was set to be 5~sec later, i.e., at $t=97.5$sec, to leave a conformable margin for accommodating the time required for path planning computations. The estimated linear and angular velocities of the target along with the estimated inertial parameters are plotted in Figs.\ref{fig:velocity} and \ref{fig:param}, respectively. The motion planner progressively updates the robot trajectories based on the most recent state and parameter estimation until the vision system fails, after which the state/parameter estimation is no longer updated from the faulty vision data. Fig.~\ref{fig:icp_eps} displays the time-histories of the point-matching error of the 3D registration along with the predicted position of the grasping fixture. The graphs reveal that obstruction of the vision sensor by the approaching servicing manipulator occurs about 10~sec prior to completion of the pre-capturing phase. This event results in the metric fit error increasing to such an extent that the fault-detection logic renders the estimator gain zero. However, the plots in the figure demonstrate that the estimator still provides a reliable prediction of the target position after the vision obstruction. Figs.~\ref{fig:capture_position} and \ref{fig:capture_velocity} illustrate the position and velocity trajectories of the grasping fixture relative to the end-effector during the pre-capturing and post-capturing phases, respectively, while the trajectory of the LOS angle is shown in Fig.~\ref{fig:LOS}. The plots clearly demonstrate that the robot successfully captured the grasping fixture on the moving target at $t=121.4~\text{sec}$ and subsequently stabilized its linear and angular motions at $t=144.3~\text{sec}$. The graphs exhibit a smooth capture with both the end-effector and grasping fixture reaching the interception point with the same velocity. Furthermore, the line-of-sight (LOS) to the target grasping point remains unobstructed during the robotic capture.
The plots also reveal that the post-capturing manoeuvre of the robot simultaneously damp out the translational and rotational motion of the satellite within 22.9~sec, while respecting the maximum acceleration capability of the servicing manipulator. Fig.~\ref{fig:capture_force} depicts the time-histories of the force and torque applied to the target by the manipulator's end-effector, which are bounded according to their limits. The magnitude of the exerted force and torque are also illustrated by dotted lines in the figure, demonstrating that the forces and torques are indeed saturated. The plots clearly indicate that the robot successfully captured the grasping fixture on the moving target at $t=121.4~\text{sec}$ and stabilized its linear and angular motions at $t=144.3~\text{sec}$. The capture was smooth, with both the end-effector and grasping fixture reaching the interception point with the same velocity. The post-capture robot manoeuvres effectively dampened both the translational and rotational motion of the satellite within 22.9~sec, while respecting the specified maximum acceleration capability of the servicing manipulator. Fig.~\ref{fig:capture_force} shows the time-histories of the force and torque exerted by the manipulator's end-effector on the target, which are bounded by their limits. The magnitude of the exerted force and torque are also indicated by dotted lines, demonstrating that they are saturated.

\begin{figure}
\centering{\includegraphics[clip,width=8cm]{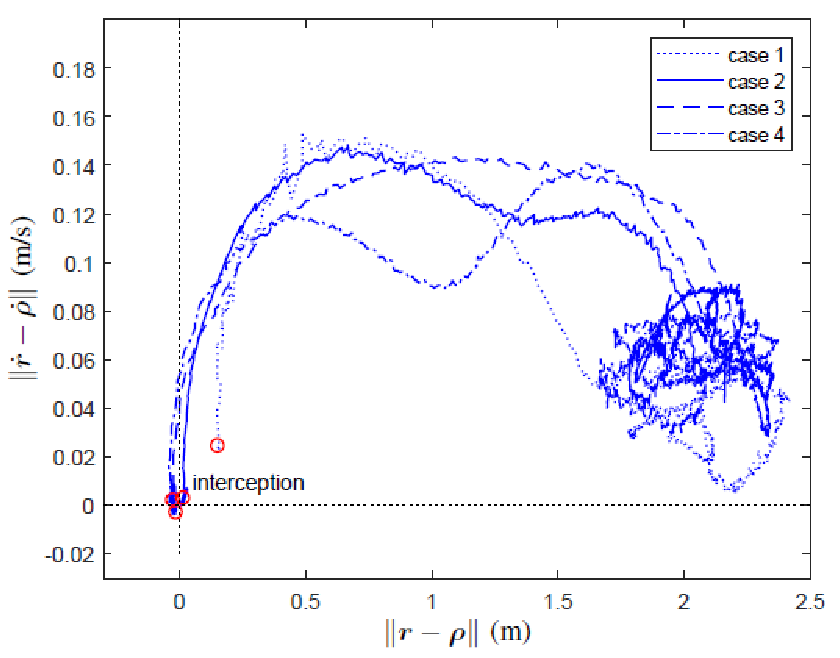}} \caption{Trajectories of relative position versus relative velocity for different test cases.}\label{fig:rel_trajectory}
\end{figure}

In summary, the optimal trajectory planning and control scheme enables the manipulator to capture and stabilize the target as quickly as possible within a total time of 46.9~sec, given the manipulator capabilities. Fig.~\ref{fig:rel_trajectory} illustrates trajectories of the distance between the end-effector and the target versus the relative velocities from multiple experimental results. Test case 1 corresponds to the motion estimation and control scheme without incorporation of the fault-detection logic, while test cases 2, 3, and 4 correspond to motion estimation and control with incorporation of the fault-detection logic under different initial and operational conditions. In all test cases, the vision system fails when the manipulator is close enough to the target and thus inevitably obstructs the field-of-view of the camera, which is placed at a fixed point in the workspace. The plots clearly demonstrate that the motion estimation and control scheme without incorporation of the fault-detection logic did not succeed in achieving the basic objective of rendezvous \& capture due to the large rendezvous position and velocity errors. However, the proposed motion estimation and control scheme achieved an average rendezvous position and velocity errors of about 2.6~cm, which is lower than the 4~cm capture envelope of the robotic gripper hand. Therefore, successful rendezvous \& capture of the target becomes possible in spite of the occlusion.

\section{Conclusion}
We have presented an integrated vision-guidance and optimal control method for autonomously capturing and stabilizing a tumbling and drifting target object in a time-critical manner. The method could take into account various operational and physical constraints, including ensuring a smooth capture, handling line-of-sight (LOS) obstructions of the target, and staying within the acceleration, force, and torque limits of the robot. The integrated system achieved not only a seamless transition and real-time switching between control systems but also self-tuning of both controllers through the processing of visual data. By incorporating a fault detection logic based on metric fit of the registration algorithm and prediction error, we were able to implement a fault-detection and recovery strategy, ensuring continuous visual feedback even in the event of obstruction of the vision sensor.
We successfully implemented and tested the vision-guided control scheme on the CSA satellite simulator testbed, which featured two manipulator arms simulating the motions of a tumbling satellite and a servicing robot. The experimental results demonstrated successful execution of capturing and stabilizing the tumbling and drifting satellite through sequential pre- and post-capturing operations, despite the presence of multiple operational constraints and obstructed 3D vision data.

\appendix

The linearized process dynamics is described by
\begin{subequations} \label{eq:dot_deltax}
\begin{equation}
\delta \dot{\bm x} = \bm F \delta \bm x + \bm L \bm w,
\end{equation}
\begin{align} \label{Eq:F}
\bm F &=
\begin{bmatrix} - [\hat{\bm\omega} \times] & \frac{1}{2} \bm I & \bm 0 & \bm 0 & \bm 0 & \bm 0 & \bm 0 \\
\bm 0& \left(\frac{\partial\bm\phi}{\partial\bm\omega}\right)_{\!\! \hat{\bm x}}   & \bm 0 & \bm 0 & \left(\frac{\partial\bm\phi}{\partial\bm\sigma}\right)_{\!\!
\hat{\bm x}}  & \bm 0 & \bm 0 \\
\bm 0 & \bm 0& \bm 0 & \bm I & \bm 0 & \bm 0 & \bm 0   \\
\bm 0 & \bm 0 & \bm 0 & \bm 0 & \bm 0 & \bm 0 & \bm 0
\end{bmatrix} \\ \label{Eq:G}
\frac{\partial\bm\phi}{\partial\bm\sigma} & = \begin{bmatrix} \omega_y \omega_z & 0 \\ 0 & \omega_x \omega_z  \\ \frac{\sigma_2^2 -1}{(1+ \sigma_1 \sigma_2)^2}
\omega_x \omega_y & \frac{\sigma_1^2 -1}{(1+ \sigma_1  \sigma_2)^2} \omega_x \omega_y \end{bmatrix}.
\end{align}
\end{subequations}
The equivalent discrete-time system of system \eqref{eq:dot_deltax} is
\begin{equation} \label{eq:discere_sys}
\delta \bm x_{k+1} = \bm\Phi_k \delta \bm x_k + \bm w_k.
\end{equation}
Here, $\bm w_k$ is discrete-time process noise, $t_{\Delta} = t_{k+1}-t_k$ is the sample time, and the state transition matrix is denoted by $\bm\Phi_k = \bm\Phi(t_k, t_{\Delta})$ where
\begin{equation} \label{eq:Phi}
\bm\Phi(t_k, t_{\Delta}) = e^{\bm F(t_k) t_{\Delta}} \approx \bm I +  t_{\Delta} \bm F(t_k).
\end{equation}
The covariance of process noise associated with the discrete-time systems  $\bm Q_k=E[\bm w_k \bm w_k^T]$ can be obtained from
\begin{equation} \notag
\bm Q_k= \int_{t_k}^{t_k+t_{\Delta}}
\bm\Phi(t_k,\tau) \bm L \; \mbox{diag}\big( \sigma_{\tau}^2 \bm I, \sigma_{f}^2 \bm I  \big) \; \bm L^T \bm\Phi^T(t_k,\tau) {\rm d}
\tau,
\end{equation}
Using \eqref{Eq:F}, \eqref{Eq:G}, and \eqref{eq:Phi} in the above integral, we get
\begin{equation} \label{eq:Qk}
\bm Q_k= \begin{bmatrix} \bm Q_{11} \sigma_{\tau}^2 &  \bm Q_{12} \sigma_{\tau}^2 & \bm 0 & \bm 0 & \bm 0\\
\times & \bm Q_{22} \sigma_{\tau}^2 & \bm 0 & \bm 0 & \bm 0\\
\bm 0 & \bm 0 &   \frac{t_{\Delta}^3}{3}\sigma_{f}^2 \bm I &  \frac{t_{\Delta}^2}{2}\sigma_{f}^2 \bm I & \bm 0\\
\bm 0 & \bm 0 & \times & t_{\Delta}\sigma_{f}^2 \bm I & \bm 0\\
\bm 0 & \bm 0 & \bm 0 & \bm 0 & \bm 0
\end{bmatrix},
\end{equation}
where
\begin{align*}
\bm Q_{11} & = \frac{t_{\Delta}^3}{12}
\bm B^2, \quad 
\bm Q_{12}  = \frac{t_{\Delta}^3}{6} \bm B^2  \left(\frac{\partial\bm\phi^T}{\partial\bm\omega}\right)_{\!\! \hat{\bm x}}   + \frac{t_{\Delta}^2}{4} \bm B^2 \\
\bm Q_{22} & = \frac{t_{\Delta}^3}{3}  \left(\frac{\partial\bm\phi}{\partial\bm\omega}\right)_{\!\! \hat{\bm x}}    \bm B^2
\left(\frac{\partial\bm\phi^T}{\partial\bm\omega}\right)_{\!\! \hat{\bm x}}  \\ & + \frac{t_{\Delta}^2 }{2} \Big(\bm B^2  \left(\frac{\partial\bm\phi^T
}{\partial\bm\omega}\right)_{\!\! \hat{\bm x}}   +
 \left(\frac{\partial\bm\phi}{\partial\bm\omega}\right)_{\!\! \hat{\bm x}}   \bm B^2 \Big) + \bm B^2 t_{\Delta}
\end{align*}

In the other hand, defining quaternion variations $\delta \bm\mu= \bm\mu \otimes \hat{\bm\mu}^{-1}$, we can readily establish the relationship between the measured quaternion and its
variation through the following identity
\begin{equation} \label{eq:eta}
\bm\eta = \delta \bm\mu \otimes \hat{\bm\eta} \otimes \delta \bm q
\end{equation}
where $\hat{\bm\eta} = \hat{\bm\mu} \otimes \hat{\bm q}$. Then, by virtue of \eqref{eq:eta}, the observation equation \eqref{eq:h(x)} can be also realized as a nonlinear function of the state variation $\delta \bm x$, i.e.,
\begin{equation} \label{eq:h_nonlin}
\bm h (\delta \bm x) = \begin{bmatrix} \bm\rho_o + \bm A(\delta \bm q \otimes \hat{\bm q}) \bm\varrho \\ \mbox{vec} \big( \delta \bm\mu \otimes \hat{\bm\eta} \otimes
\delta \bm q \big) \end{bmatrix}.
\end{equation}
Finally, one can derive the observation sensitivity
matrix in the following form
\begin{equation} \notag
\bm H = \left( \frac{\partial \bm h}{\partial \delta \bm x} \right)_{\! \hat{\bm x}} =
\begin{bmatrix} \frac{\partial \bm\rho}{\partial \delta \bm q_v }  &\bm 0
& \bm I & \bm 0  & \bm 0 & \bm A(\hat{\bm q}) &\bm 0\\
 \frac{\partial \bm\eta_v}{\partial \delta \bm q_v }   &\bm 0  &\bm 0 &\bm 0  & \bm 0& \bm 0 & \frac{\partial \bm\eta_v}{\partial \delta \bm\mu_v}
\end{bmatrix}_{\! \hat{\bm x}},
\end{equation}
where
\begin{align*}
\frac{\partial \bm\rho}{\partial \delta \bm q_v } & = -2\bm A({\bm q})[{\bm \varrho} \times] \\
\frac{\partial \bm\eta_v}{\partial \delta \bm q_v }  & = [(\delta \bm\mu_v \times \hat{\bm\eta}_v) \times] - \delta \bm\mu_v \hat{\bm\eta}_v^T - \hat{\eta}_o(\bm I +
[\delta\bm q_v \times]) - [\hat{\bm\eta}_v \times] \\
\frac{\partial \hat{\bm\eta}_v}{\partial \delta \bm\mu_v}  &= [\delta \bm q_v \times][\hat{\bm\eta}_v \times] - \delta\bm q_v \hat{\bm\eta}_v^T + \eta_o(\bm I + [\delta
\bm q_v \times]) + [\hat{\bm\eta}_v \times]
\end{align*}
Define {\em a prioir} and {\em a posteriori} estimation errors $\delta \tilde{\bm x}_k^-= \delta \bm x_k - \delta \hat{\bm x}_k^-$ and $\delta \tilde{\bm x}_k^+=\delta
\bm x_k - \delta  \hat{\bm x}_k^+$ with associated covariances  $\bm P_k^-=E[\delta \tilde{\bm x}_k^- \delta \tilde{\bm x}_k^{-T}]$ and $\bm P_k^+=E[\delta \tilde{\bm
x}_k^+ \delta \tilde{\bm x}_k^{+T}]$. The Kalman filter gain  minimizes the performance index $E(\| \delta \tilde{\bm x}_k^+ \|^2)=\mbox{tr}(\bm P_k^+)$ subject to the
state constraints \eqref{eq:Dsigma<1}. Therefore, according to the Joseph formula, the constrained Kalman filter is the solution of the following optimization
programming
\begin{align*}
\min_{\bm K_k}  \mbox{tr} & \big((\bm I - \bm K_k \bm H_k) \bm P_k^- (\bm I + \bm K_k \bm H_k)^T + \bm K_k \bm R_k \bm K_k^T  \big) \\
\mbox{subject to:} & \quad -\bm 1 < \bm\hat{\bm\sigma}_k^+ < \bm 1
\end{align*}
The gain projection technique can be applied to impose the inequality constraints for the estimation process \cite{Teixeira-Chandrasekar-2008}. In this method if  the unconstrained {\em a posteriori} estimation $\delta \hat{\bm x}_k^+$ does not satisfy the inequality constraints,
then  the state estimation is projected to the constraint boundary in the direction of {\em a prioir} estimation. This effectively modified the Kalman gain as follows:
\begin{equation}
\bm K_k = \bm\Lambda_k \bm K^u_k,
\end{equation}
where $\bm\Lambda_k$ was previously defined in \eqref{eq:Lambda_k} and $\bm K^u$ is the unconstrained Kalman gain given by
\begin{equation} \label{eq:Sk}
\bm K_k^u  = {\bm P}_k^- \bm H_k^T (\bm H_k \bm P_k^- \bm H_k^T + \bm R_k)^{-1},
\end{equation}
where ${\bm P}_k^+ = \big( \bm I - \bm K_k \bm H_k  \big) {\bm P}_k^-$. Subsequently the propagations of the state and covariance matrix are obtained from
\begin{subequations} \label{eq:KF_propagate}
\begin{align}\label{eq:state-prop}
\hat{\bm x}_{k+1}^- & = \hat{\bm x}_{k}^+ +
\int_{t_k}^{t_{k}+t_{\Delta}} \bm f(\bm x)\,{\text d} t\\ \label{eq:cov-prop}
{\bm P}_{k+1}^-&= \bm\Phi_{k} \bm P_{k}^+ {\bm\Phi}_{k}^T + \bm Q_{k}
\end{align}
\end{subequations}

\bibliographystyle{IEEEtran}


\end{document}